\documentclass[review]{elsarticle}

\usepackage{amssymb}
\usepackage{amsmath}
\usepackage{booktabs}
\usepackage{multirow}
\usepackage{amsfonts}
\usepackage{graphicx}
\usepackage{duckuments}
\usepackage{makecell}

\usepackage{breakurl}
\usepackage[breaklinks]{hyperref}

\usepackage{soul}
\usepackage{xcolor}
\usepackage{lineno}
\usepackage{float}

\newcommand{\dslr}{\texttt{\textbf{MOVES}}}
\newcommand{\dslruda}{\texttt{\textbf{MOVES-MMD}}}
\newcommand{\CacciaAE}{\texttt{\textbf{Caccia-AE}}}
 
\newcommand{\CacciaGAN}{\texttt{\textbf{Caccia-GAN}}}
\newcommand{\Coarse}{\texttt{\textbf{Coarse-Net}}}
\newcommand{\ds}{\texttt{\textbf{DSLR}}}

\newcommand{\siv}{\texttt{\textbf{ShapeInversion}}}

\newcommand{\subf}[2]{%
  {\small\begin{tabular}[t]{@{}c@{}}
  #1\\#2
  \end{tabular}}%
}

\begin{document}

\begin{frontmatter}

\title{MOVES: Movable and Moving LiDAR Scene Segmentation in Label-Free settings using Static Reconstruction}

\author{Prashant Kumar\footnote[1]{Corresponding Author}$^a$, Dhruv Makwana$^c$, Onkar Susladkar$^c$, Anurag Mittal\footnote[2]{Work supervised while visiting IIT Delhi, India. Permanent Address is IIT, Madras-600036, India.}$^b$, Prem Kumar Kalra$^b$,\\
$^a$School of IT, Indian Institute of Technology, Delhi-110016, India $^b$Department of CSE, Indian Institute of Technology, Delhi-110016, India\\$^c$Independent Researcher, India\\
prashant.kumar@sit.iitd.ac.in, dmakwana503@gmail.com, onkarsus13@gmail.com, amittal@cse.iitm.ac.in, pkalra@cse.iitd.ac.in}

\begin{abstract}

Accurate static structure reconstruction and segmentation of non-stationary objects is of vital importance for autonomous navigation applications. These applications \textit{assume} a LiDAR scan to consist of only static structures. In the real world however, LiDAR scans consist of non-stationary dynamic structures - moving and \textit{movable} objects. Current solutions use segmentation information to isolate and remove moving structures from LiDAR scan. This strategy fails in several important use-cases where segmentation information is not available. In such scenarios, moving objects and objects with high uncertainty in their motion \textit{i.e. movable objects}, may escape detection. This violates the above assumption. We present \dslr{}, a novel GAN based adversarial model that segments out moving as well as movable objects in the absence of segmentation information. We achieve this by accurately transforming a dynamic LiDAR scan to its corresponding static scan. This is obtained by replacing dynamic objects and corresponding occlusions with static structures which were occluded by dynamic objects. We leverage corresponding static-dynamic LiDAR pairs. We design a novel discriminator, coupled with a contrastive loss on a smartly selected LiDAR scan triplet. For datasets lacking paired information, we propose \dslruda{} that integrates Unsupervised Domain Adaptation into the network. We perform rigorous experiments to demonstrate state of the art dynamic to static translation performance on a sparse real world industrial dataset, an urban and a simulated dataset. \dslr{} also segments out movable and moving objects without using segmentation information. Without utilizing segmentation labels, \dslr{} performs better than segmentation based navigation baseline in \textit{highly dynamic} and \textit{long} LiDAR sequences. 
\end{abstract}

\begin{keyword}
Label-free LiDAR Segmentation, Static LiDAR Reconstruction, Dynamic Environments.   
\end{keyword}

\end{frontmatter}

\section{Introduction}
\label{introduction}

\begin{figure}[htbp]
\centering
\includegraphics[width=0.6\linewidth, height=0.4\linewidth]{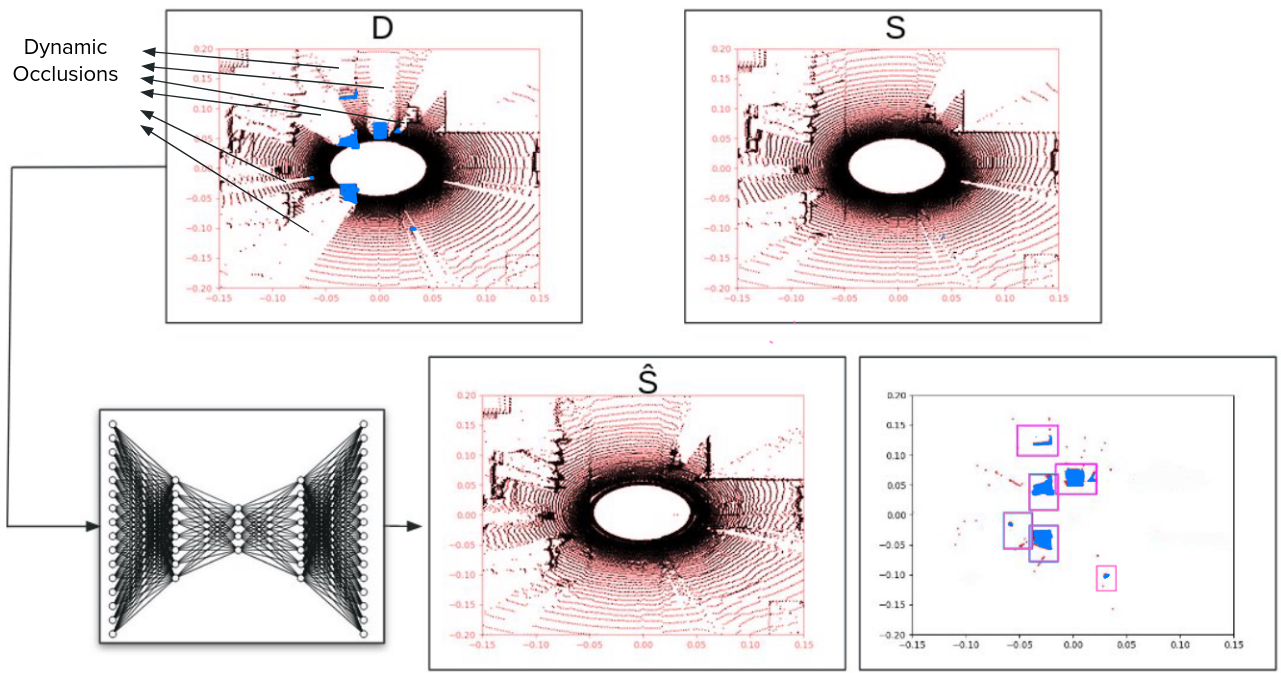}  
\caption{\textbf{Top Row}: A corresponding dynamic-static LiDAR pair $(D,S)$. Left - Dynamic scan, $D$ which has dynamic objects (in \textcolor{blue}{blue}) and dynamic occlusions (indicated by arrows). Right - corresponding static scan, $S$ in the same environment. \textbf{Bottom Row}: Our model in action. Given the dynamic scan $D$ as input, our model generates a corresponding static scan $(\hat{S})$ (middle), that approximates $S$. We subtract $D$ from $\hat{S}$  to extract moving and movable objects (pink boxes in rightmost image). }
\label{fig:movable-first}
\end{figure}

Handling non-stationary objects is an important problem with major applications for LiDAR (Light Detection and Ranging) based Autonomous Vehicles (AVs). Current solutions extract existing static structures by segmenting and removing moving objects  (\citet{chen2021moving}) with the help of segmentation and object information. They use this information to train a model to identify and remove moving objects in real time. Collecting segmentation labels is laborious, expensive and time-consuming, especially for several popular and important settings like industrial environments, fulfilment centers, urban settings, etc. In addition, common everyday scenarios also involve \textit{movable} objects. They have the capacity to move but may be stationary at a given instant, and can escape detection in segmentation label free settings due to uncertain motion. When moving, these objects add to the moving parts of a scene, thereby leading to erroneous pose estimates.  Thus, they are a potential hazard to AVs.

We address the above issues by proposing \dslr{}, a GAN based generative network that enables precise segmentation of dynamic - movable as well as moving objects. This is achieved in the absence of any form of segmentation information, by generating corresponding static reconstructions for a given dynamic LiDAR scan (Figure \ref{fig:movable-first}). \dslr{} is trained with corresponding static and dynamic LiDAR scan pairs which are easier to collect and are time and cost effective (\citet{dslr-git}) as compared to labelled information collection. In \dslr{}, we introduce an intuitively designed discriminator that enforces an implicit attention to dynamic objects. This is coupled with an intuitively designed contrastive strategy that reinforces the dissimilarity between dynamic and static scans. 

We list the contributions of our work here:
\begin{itemize}

    \item We present \dslr{}, a GAN-based adversarial model for \textit{generating static background} of occluded parts of a dynamic LiDAR scan (dynamic to static translation) and henceforth \textit{segmenting out moving and movable  parts of scenes}. This is achieved in a segmentation-label devoid setting. \dslr{} uses a novel couple discriminator to discriminate carefully chosen pairs of static and dynamic input. This design choice helps our model to explicitly focus its attention on moving as well as movable objects, without external supervision.
 
    \item To reinforce the attention of our couple discriminator on the pattern of moving and movable objects, we design a novel contrastive strategy. We utilize static-dynamic LiDAR correspondence to design hard negative triplets of scans for contrastive learning. 
    
    \item We quantitatively demonstrate the superior quality of static reconstructions on 3 varied datasets. We qualitatively demonstrate the usefulness of \dslr{} in segmenting out movable as well as moving objects without using segmentation labels. 
       \item To work well with datasets lacking static-dynamic correspondence, we propose \dslruda{}. We use Unsupervised Domain Adaptation to adapt models trained on correspondence datasets to datasets lacking correspondence. Despite working in segmentation devoid settings, \dslruda{} performs comparable to or better than segmentation based baselines for navigation in long sequences with \textit{highly dynamic environments}.
\end{itemize}
\section{Related Work}
\label{related}
 Generative modeling for LiDAR has seen considerable development with the advent of several pattern recognition systems - diffusion models, generative models, etc. They use LiDAR point cloud or range image representation for underlying pattern recognition and modelling of LiDAR. Generating occluded static parts of scenes has been explored for several modalities: images, point clouds and recently LiDAR. 
 via synthetic-to-real transfer. 
 (\citet{wang2022coarse}) propose a coarse-to-fine network for dynamic-to-static image translation, without any segmentation information and show impressive results. 
 (\citet{ahn2023unsupervised}) propose to inpaint a dynamic semantic segmentation mask by replacing dynamic objects based labels with static object labels.
  For point clouds, (\citet{fei2023dctr}) address the problem of unstructuerd point prediction in lcoal patches and the inefficient utilization of local features. They propose a dual channel transformer that leverages point and channel-wise attention to generate locally compact and structured point clouds. (\citet{li2023progressive}) model the point cloud generation process a two-level hierarchy of distributions to learn the pattern of the point clouds. They propose a hierarchical consistency variational autoencoder that ensures that  the fine-grained parts and underlying shape and can be reconstructed with high precision. \citet{zhang2023learning} alleviate distortion of geometric details by formulating the  point cloud completion as a geometric transformation problem. They exploit repetitive geometric structures in 3D shapes to generate complete shapes.
  (\citet{zhang2021unsupervised}) uses GAN inversion for point cloud completion and shows remarkable results on standard shape datasets. (\citet{kumar2021dynamic}) achieves static translations for LiDAR scans using an adversarially trained model. They implicitly do not use segmentation information but have a variant that uses it for static translation. (\citet{yang2021semantic}) approach scene completion for LiDAR with the assistance of semantic-segmentation. (\citet{rist2021semantic}) attempt semantic scene completion using local Deep Implicit Functions as a learning based method. Their method focuses a continuous scene pattern representation unlike existing voxelization based works - this helps to encoder accurate semantic and geometric properties. (\citet{xia2023scpnet}) also perform semantic scene completion using a novel knowledge distillation objective - Dense-to-Sparse Knowledge Distillation.

  For LiDAR, (\citet{zyrianov2022learning}) introduce LiDARGen that generates realistic high quality LiDAR samples using a stochastic denoising process that leverages the score-matching energy based model. They suffer from a major drawback - a high sampling rate of 20s/scan. This is infeasible for real world navigaiton that require a new scan within 100ms - freuqnecy of 10 Hz. (\citet{zhang2023nerf}) introduced NeRF-LiDAR, a NeRF based representation of scenes and use them to generate LiDAR scans. For a given LiDAR scene, NeRF-LiDAR requires multi-view images and semantic segmentation labels , both of which are unavailable in our setting. (\citet{xiong2023learning}) present UltraLiDAR that learns a discrete pattern of LiDAR. For LiDAR completion, the authors rely on explicit intervention to identify the codes for separate objects which have to be perturbed to obtain the resultant LiDAR.

Most static translation methods for point clouds discussed above focus on enhancing the power of the generator. Our focus is to enable the discriminator to learn precise discriminatory features between static and dynamic LIDAR scans - moving and movable objects.

\textbf{Contrastive Learning:}
Contrastive Learning focus on learning rich patterns in data by grouping samples with the same labels together while distancing out dissimilar ones.
Some well-known contrastive losses include Maximum Margin Contrastive Loss (\citet{hadsell2006dimensionality}), Triplet Loss (\citet{weinberger2009distance}) and multi-class N-pair loss (\citet{sohn2016improved}).
Triplet Loss operates on a triplet of latent vectors. Two vectors (the anchor and positive) share the same label while the third (negative) belongs to a different label. The objective of triplet loss is to find patterns in the dataset that keep the positive sample within an $\alpha$-neighborhood of the anchor while forcing the negative sample out of the $\alpha$-neighborhood, $\alpha$ being a margin hyperparamter.
The multi-class N-pair loss  is a generalization of the triplet loss.
Contrastive Learning with hard negatives (\citet{robinson2020contrastive}) involves selecting negative samples that are difficult to distinguish from the anchor sample. Several approaches (\citet{xie2022delving}) deploy hard negative mining for contrastive learning in order to learn better representations for downstream tasks.
\section{Problem Formulation}

    Throughout our paper, a corresponding static-dynamic LiDAR scan pair, refers to a LiDAR scan pair $(s_i, d_i)$ where both the scans share exactly the same surrounding. The only difference is that $d_i$ consists of movable and moving objects, unlike $s_i$ (Figure \ref{fig:movable-first}). We have a set of static scans $\textbf{St}=(s_i, i: 0,1,2...)$ and corresponding dynamic scans denoted as $\textbf{Dy}=(d_i, i: 0,1,2...)$. Given a dynamic scan $d_i$, and its latent representation on the dynamic manifold, ($r_{d_i}$): our goal is to map $r_{d_i}$  to its closest corresponding point ($\overline{r}_{s_i}$) on the static manifold. $\overline{r}_{s_i}$ must produce the closest corresponding static LiDAR $\overline{s}_{i}$, by generating static background of the occluded regions, as seen in $s_i$. We achieve this by developing a novel discriminator and a contrastive strategy that allows precise discrimination of dynamic objects and occlusions from static objects and occlusions. The generated static structures along with the input dynamic LiDAR are used for movable and moving object segmentation.
\section{Methodology}
\subsection{\dslr{} Formulation}
We describe \dslr{}'s architecture here - generator, discriminator, contrastive learning for the discriminator and the training setup for MOVES, below.
\subsubsection{Generator}
We use range image-based LiDAR preprocessing, which allows us to use image-based processing on LiDAR effectively. We use a conditional GAN architecture for the  static translation. We use the Encoder ($E_\phi$) and Decoder ($D_\theta$) from (\citet{caccia2018deep}) for our generator (\textit{G}). Given $d_i\in \textbf{Dy}$ as input, the goal of $G$ is to generate $\overline{s_i}$  (\textit{s.t.} $\overline{s_i}$ resembles $s_i$). 
\subsubsection{Discriminator}
\label{subsub-disc}
We provide an overview of the challenges while designing the discriminator. \textbf{(1)} Discriminating a static scan ($s_i$) from the dynamic scan ($d_i$) is challenging because both are generated from similar LiDAR distribution and share similar patterns - scenes and structure. Dynamic LiDAR scans are equivalent to static LiDAR with noise in the form of dynamic objects. Latent feature vectors of both consists of mostly homogeneous and almost indistinguishable patterns. Features and patterns that help in discrimination (i.e. dynamic objects and occlusions) are limited because they occupy a small percentage of points in the scan compared to the common static regions. This leads to an imbalance w.r.t. the discriminatory features, making them difficult to learn. \textbf{(2)} A vanilla discriminator for static scans \textit{v/s} dynamic scans, suffers from a \textit{simplicity bias} (\citet{shah2020pitfalls}) - the discriminator needs to only pay attention to the \textit{simplest features} for discrimination i.e. (moving objects), to distinguish them from static scans. It saturates prematurely by learning these simplest features. Movable objects (which are difficult to discover due to uncertain motion) do not contribute significantly to the learning because the model never needs to use these complex movable object features. This leads to  sub-optimal translations as shown in Ablation studies (using a GAN with a vanilla discriminator) (Section \ref{sec:ablation} and Table \ref{tab:ablation-disc}).

To address the above challenges, we design a novel \textit{couple discriminator} ($D$) for our proposed GAN, which operates on the latent space of the LiDAR scans. $D$ takes a pair of LiDAR latent representations as input.

Let ($r_{s_i}$, $r_{s_j}$, $r_{d_k}$) be the latent representation for the triplet (${s_i}$, ${s_j}$, ${d_k}$) where $s_i$, $s_j \in \textbf{St}$ and $d_k \in \textbf{Dy}$. We construct 2 pairs from the triplet: ($r_{s_j}$, $r_{s_i}$) which is labelled as real, and ($r_{d_k}$, $r_{s_i}$), which is labelled as fake by the \textit{D}.

In order to classify the above pairs as real and fake respectively, \textit{D} is forced to  amplify discriminatory features in dynamic scan, $d_k$. Instead of competing static representations against dynamic (as in a vanilla discriminator), the discrimination occurs between similarities in static scans in the  pair ($r_{s_j}$, $r_{s_i}$) v/s differences between static and dynamic scan in the pair ($r_{d_k}$, $r_{s_i}$). This helps $D$ to focus only on the discriminatory features between static and dynamic scans without using segmentation labels. $D$ can explicitly focus on dynamic objects and occlusions in $d_k$ while ignoring the common static regions. It is also able to differentiate between occlusions due to static objects and occlusions due to dynamic objects, thereby enabling \textit{G} to only remove dynamic occlusion and objects while retaining static occlusions and static objects. This behaviour is further assisted by the absence of moving and movable objects in static scans.

Further, to stabilize the training of $D$, we replace the negative log-likelihood objective of the discriminator with the least square objective (\citet{mao2017least}). The idea behind using Least Square GANs is two fold: first is to ensure that $D$ receives smooth and non-saturating gradients, thereby allowing $G$, which is also trained via gradients through $D$, to be trained effectively. Secondly, given gradients could be weak due to the structural and pattern similarity between static and dynamic scans, we would even want to penalize LiDAR pairs that are far from the discriminator decision boundary,  though on the correct side. This allows $G$ to generate static scans closer to the decision boundary to avoid learning saturation (\citet{mao2017least}). It helps to generate more gradients while updating $G$, which allows reconstructed samples to be closer to the real manifold of static scans.

\textbf{Contrastive Learning on  Discriminator:} Dynamic and static scans are generated from similar distributions and share similar patterns (except for movable and moving objects). Learning discriminatory features between them is non-trivial. To assist $D$ in paying attention only to the discriminatory patterns between them, we use a contrastive loss to \textit{contrast dynamic scans from static} ones while bringing static scans closer. Here we use a careful choice of static and dynamic scans using \textit{hard negative mining}. Recall the triplet above $(s_i\in St, s_j\in St, d_k\in Dy)$, used while training the discriminator. In this triplet, we now restrict \textit{$d_k$ to be the corresponding dynamic scan for $s_{i}$} ($i.e.$ $d_{k}$ is now $d_{i}$).
\begin{figure}[t]
  \centering
  \includegraphics[width=0.6\linewidth, height=0.5\linewidth]{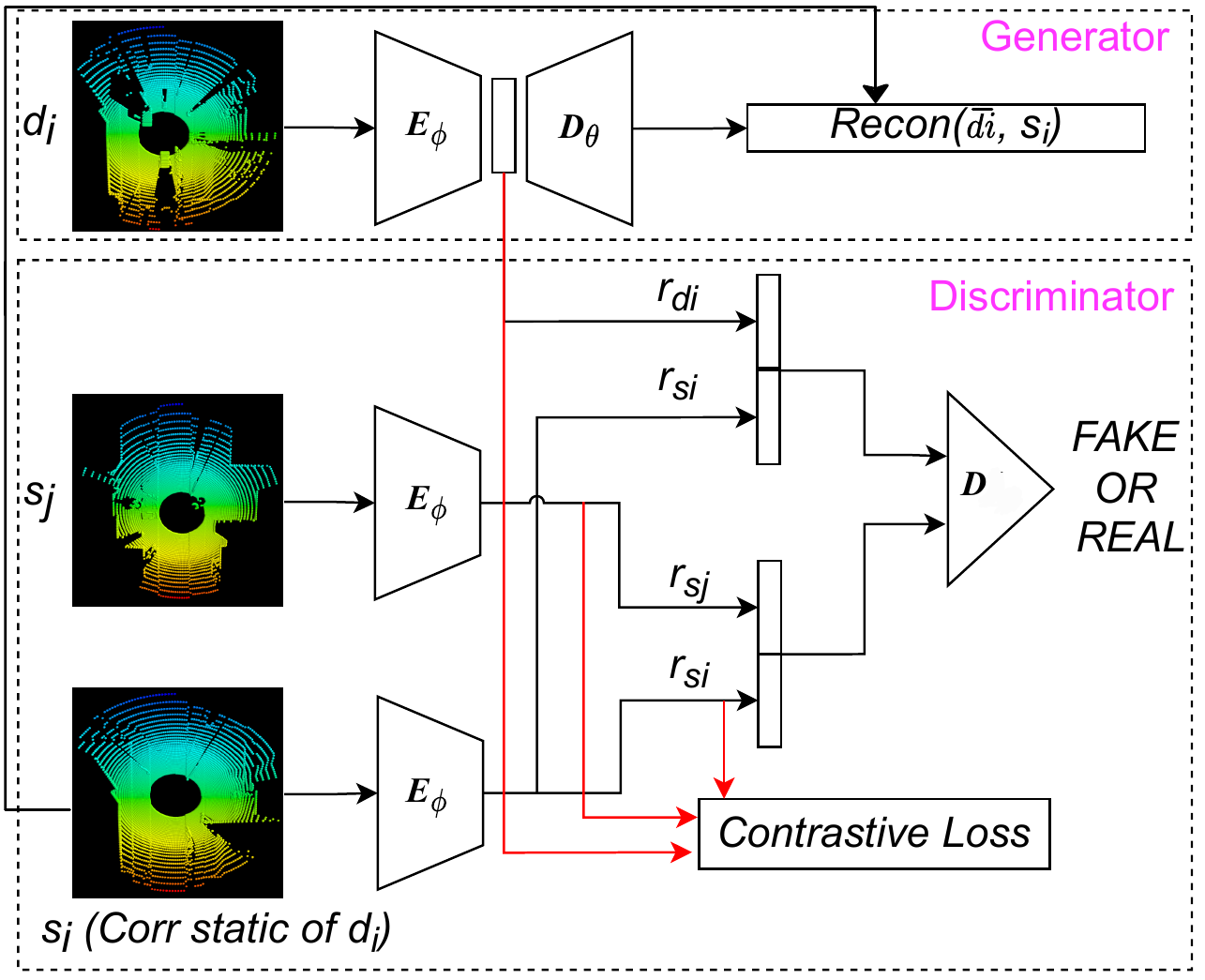}  
  \caption{Architecture Diagram for \dslr{}}
\label{fig:carlagan}
\end{figure}
We now explain the reason behind this choice. We use the notation (${s_i}$, ${s_j}$, ${d_k}$) instead of ($r_{s_i}$, $r_{s_j}$, $r_{d_k}$) for the sake of simplicity.
Consider $s_i^+$ as the anchor, $s_j^+$ as positive, and $d_i^-$ \textit{(corresponding dynamic/hard negative for $s_i^+$)} as the negative scan. In the ideal case, the anchor $s_i^+$, and the positive scan $s_j^+$ should be closer, while the anchor and the negative scan should be far away in the latent space. However due to our careful selection of the   negative sample, the anchor $s_i^+$, is more closer to the negative sample $d_i^-$, than the positive sample $s_j^+$ in the latent space. This is because the anchor $s_i^+$, and the negative sample $d_i^-$ are a corresponding pair and have very high similarity despite belonging to different classes.

The samples $s_i^+$ and $d_i^-$ are a corresponding static-dynamic pair, are perceived in the same environment and share the same static structures and patterns. Thus, they are very close in the latent space. The only difference between them - movable and moving objects, accounts for the discriminatory features. Contrasting them \textit{w.r.t.} each other forces $D$ to implicitly focus on the only available differences between them - moving and movable objects in $d_i^-$, and the corresponding static background features in static scan ($s_i^+$). This ensures that the signals that flow to $G$, explicitly focus on the above discriminatory patterns and structures, thereby helping to identify moving and movable objects and corresponding occlusions.
It also helps in differentiating \textit{occlusion due to dynamic objects from occlusions due to static objects}, thereby overcoming limitations of a vanilla discriminator as discussed in the previous sections. These improvements strengthen the discriminatory signals to $D$ while also helping $G$ focus on dynamic objects and occlusions explicitly, for reconstructing background static scenes. Given \textit{m} is a distance metric, and $\alpha$ is the neighbourhood parameter, contrastive triplet loss, $L_{tri}$ is defined as:
\begin{equation}
        =max(0, m( s_i^+, s_j^+)  - m(d_i^-, s_i^+) + \alpha)
\end{equation}

\subsection{Training Setup for Conditional Translation}
\label{setup-carla}

 \textit{Discriminator}: We use a MLP-based sequential neural network consisting of 6 layers with sigmoid non-linearity. As detailed above, for training, we consider a triplet of the form $(s_i\, s_j, d_i)$ (Figure \ref{fig:carlagan}) for training $D$. We generate latent vectors of the triplet using the encoder ($E_{\phi}$ of $G$), and create pairs ($r_{s_j}$, $r_{s_i}$) and ($r_{d_i}$, $r_{s_i}$) using the triplet. The former is labeled as real while the latter as fake by our discriminator. $D$ is trained using the discriminatory signals from the real $v/s$ fake classification and  the Contrastive triplet loss.

\textit{Generator:} The generator $G$ is fed with a dynamic scan ($d_i$) and is trained with the reconstruction loss using the corresponding static scan $s_i$ (Figure \ref{fig:carlagan}). Note that the reconstructed scan does not flow into the discriminator $D$. Its latent vector $r_{d_i}$ flows in the discriminator because we operate on the latent space. This latent vector ($r_{d_i}$) is a constituent of the pair ($r_{d_i}$, $r_{s_i}$), which is labeled as fake by $D$ (Figure \ref{fig:carlagan}).
The generator is trained using the gradients received from $D$, as well the gradient signals from the decoder($D_{\theta}$ of $G$).

To enhance the gradients to $G$ and  $D$, we employ a relativistic least square GAN for \dslr{}.
    The generator $G(\phi,\theta{})$ takes $d_i\in Dy$ as input and generates $\overline{s_i} \sim s_i$ (where ${s_i} \in St$  is the corresponding static of $d_i$). On the other hand, given a triplet of LiDAR scan latent representations $(r_{s_i}, r_{s_j}, r_{d_i})$, the couple discriminator ($D$) classifies the couple $(r_{s_j}, r_{s_i})$ as real and $(r_{d_i}, r_{s_i})$ as fake. 

     We list the loss formulation for the Discriminator ($Loss_{D}$) as 
    \begin{equation}
    \begin{split}
           &= D((r_{s_j},r_{s_i}), real) + D((r_{d_i},r_{s_i}), fake)\\ &+ {L_{tri}}(r_{s_i}, r_{s_j}, r_{d_i})
    \end{split}
    \label{eq:disc}
    \end{equation}
    and for the generator ($Loss_{G}$) as 
    \begin{equation} 
    \begin{split}   
    &= D((r_{d_i},r_{s_i}), real) +Recon(\overline{d_i},s_i)
    \end{split}
    \label{eq:gen}
    \end{equation}
    where \textit{Recon}() denotes the Reconstruction Error while $L_{tri}$ denote the Contrastive triplet loss and $\overline{d_i}$ is the reconstructed output of the $G$, given $d_i$ as input.

    Assuming $P(St)$ denotes the distribution for the static LiDAR scans, while $P(Dy)$ denotes the distribution for the dynamic LiDAR scans. \\

    Equation \ref{eq:disc} can be expanded as
    \begin{equation}
    \begin{split}
     & =
    {\mathbb{E}}_{s_i,s_j\sim P(St)}[(DI(r_{s_j},r_{s_i}) -1)^{2}]\\
    & + {\mathbb{E}}_{d_i\sim P(Dy),s_i\sim P(St)}[(DI(r_{d_i},r_{s_i}) - 0)^{2}] \\ 
    & + {L_{tri}}(r_{s_i}, r_{s_j}, r_{d_i})
    \end{split}
    \end{equation}
    while equation \ref{eq:gen} can be expanded as
    \begin{equation}
    \begin{split}
         =
         {\mathbb{E}}_{d_i\sim P(Dy),s_i\sim P(St)}[(DI(r_{d_i},r_{s_i}) -1)^{2}] +
    Recon(\overline{d_i}, s_i)
    % }
    \end{split}
    \label{egenloss}
    \end{equation}

Now we explain the training process of \dslr{}.
Given a batch of triplets $(s_i,s_j,d_i)$, we use Encoder ($E_{\phi}$) of $G$, to extract the latent representations, i.e. $(r_{s_i},r_{s_j},r_{d_i})$. 
We create the real and fake pairs, ($r_{s_j}$, $r_{s_i}$) and ($r_{d_i}$, $r_{s_i}$) respectively. $D$ it is trained to classify these pairs correctly during training. $G$ is trained
% given a dynamic scan ($d_i$) as input, 
to produces a latent representation ($r_{d_i}$), such that the pair ($r_{d_i}$, $r_{s_i}$) is labeled as real instead of fake by $D$. In order to achieve this, \textit{G} must ensure that for the given dynamic scan $d_i$, it must produces a latent representations that is close to the corresponding static LiDAR scan $r_{s_i}$, on the static manifold. This allows us to achieve our translation.
 
\begin{figure}[t]
  \centering
  \includegraphics[width=0.6\linewidth, height=0.6\linewidth]{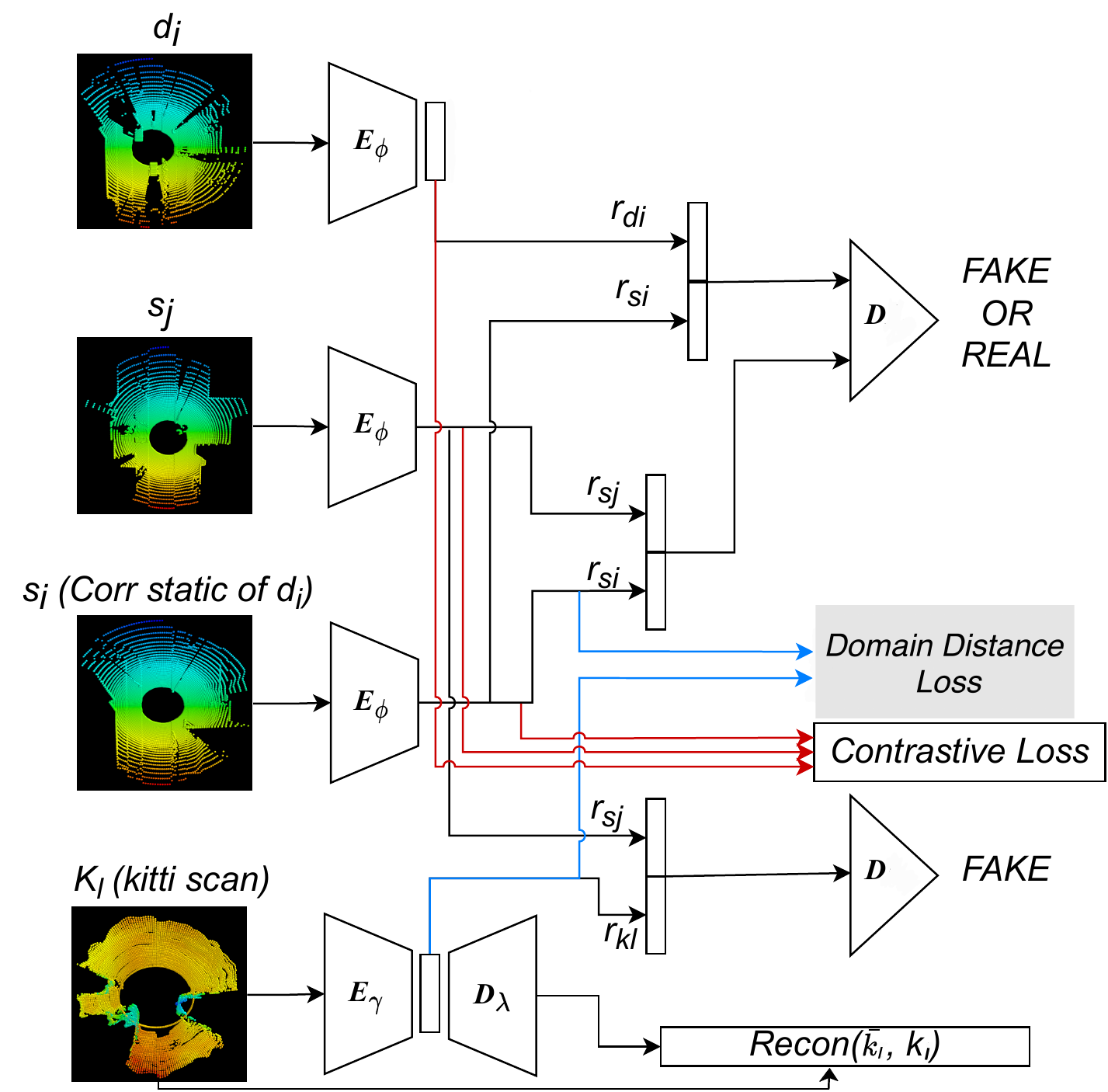}  
  \caption{Architecture Diagram for \dslruda{}}
\label{fig:ARCHDAG}
\end{figure}
\subsection{\dslruda{} }
A challenge for our approach is the unavailability of corresponding ground truth static for dynamic LiDAR scans for some datasets. To overcome this, we resort to Domain Adaptation and  follow (\citet{Borgwardt2006IntegratingSB}) to integrate a domain discrepancy loss ($L_{domain\_distance}$) between the paired and the unpaired domain in the latent space. There exists several other methods and loss functions that can be used here. We use (\citet{Borgwardt2006IntegratingSB}) for adaptation because it is straight forward, requires minimal change to integrate in \dslr{} and works well. It projects the data points from both domains in the reproducible Kernel Hilbert space followed by computing the average difference between them. For our purpose the source domain is CARLA-64 dataset (\citet{dosovitskiy2017carla}) and the target domain is KITTI dataset (\citet{geiger2012we}). We consider the same architecture of \dslr{} for \dslruda{} with a few additions (Figure \ref{fig:ARCHDAG}) -
\\ \textbf{(a)} We introduce a new generator for KITTI - $G_{K}(E_{\gamma}, D_{\lambda})$. KITTI scans are labelled as $k_l$.
\\ \textbf{(b)} Latent representations of KITTI scan ($E_{\gamma}(k_l)$ or $r_{k_l}$) is paired with $r_{s_j}$ :  ($r_{k_l}, r_{s_j}$). It is labelled as fake by $D$. \\
 \textbf{(c)} Domain distance loss is calculated between $r_{k_l}$ (latent vector of dynamic scan in target domain) and $r_{s_i}$ (latent vector of static scan in source domain).\\ 
 \textbf{(d)} KITTI dynamic scan is reconstructed using the KITTI generator $G_{K}(E_{\gamma}, D_{\lambda})$. It helps to retain the static structures of the target scan. 

We use a triplet of latent representations obtained from LiDAR scans in the source CARLA-64 domain $(r_{s_i}, r_{s_j}, r_{d_i})$ as discussed in Section \ref{setup-carla}. We use another latent representation derived from the target KITTI domain, $r_{k_l}$  using the KITTI generator $G_{K}(E_\gamma,D_\lambda)$.

    Discriminator Loss for \dslruda{}, $Loss_{D-kitti}$ is as follows:
    \begin{equation}
    \begin{split}
           &= Loss_{D} +  D((r_{k_l}, r_{s_j}), fake)  
    \end{split}
    \end{equation}  
    \begin{equation}
        \begin{split}        
           =Loss_{D} +  {\mathbb{E}}_{k_l\sim P(kitti),s_j\sim P(St)}[(D(r_{k_l},r_{s_j}) - 0)^{2}] 
    \end{split}
    % \label{eq:disc}
    \end{equation}
    and the generator loss $Loss_{G-kitti}$ is as follows:
    \begin{equation}
    \begin{split}
           &= Loss_{G} +  Recon(\overline{k_l}, k_l)
           +L_{Domain-distance}(r_{k_l},r_{s_j})
    \end{split}
    % \label{eq:disc}
    \end{equation}

\section{Experiments}
We evaluate \dslr{} on 3 diverse datasets including an industrial and an urban dataset using 5 baselines that show strong modeling performance on LiDAR for static reconstruction. We show quantitative results depicting the accuracy of the reconstruction in Table \ref{tab:carlaresults} and qualitative results for movable and moving object segmentation in Figure \ref{fig:movable-first}, \ref{fig:ati-moving-movable} and \ref{fig:movable-second}.
We also show the benefit of predicted static structures on navigation in Table \ref{tab:slamresults}.

\subsection{Datasets}
\textbf{ARD-16}: ARD-16 is a real-world industrial  LiDAR dataset collected using a 16-beam VLP-Puck LiDAR sensor. It has paired correspondence information available. This dataset is challenging because it is 4$\times$ sparse compared to the other datasets. It has very few points falling on static structures as well as on dynamic objects. This makes it challenging to learn the features of dynamic and static objects. For training and testing \dslr{} on ARD-16, we follow the protocol mentioned at (\citet{dslr-git}).\\
\textbf{KITTI}: We use KITTI (\citet{geiger2012we}) dataset to demonstrate that our model works well in the absence of correspondence information. For KITTI, less than 7\% of the available scans are static scans without any correspondence information available. Therefore, we are not able to use standard metrics like Chamfer's Distance for static translation evaluation. We resort to LiDAR Quality Index (LQI) which positively correlates with Chamfer's Distance and has been used in such scenarios (\citet{kumar2021dynamic}). More details in Section \ref{metrics}. \\
 \textbf{CARLA-64} : CARLA-64 is extensive dataset collected using CARLA simulator (\citet{Dosovitskiy17}). It is collected using the settings of VLP-64 Velodyne LiDAR on a virtual city map. One of the motivation behind using CARLA is the availability of paired correspondence between static and dynamic scans. Such paired dataset collection has been shown to be viable and easy in urban as well as industrial settings at a fraction of the cost, time and man power required for the annotation of segmentation information (\citet{kumar2021dynamic}). We perform our experiments on the latest version of the dataset and follow the train/test protocol mentioned at (\citet{dslr-git}). 
 
 All the three datasets have significant pattern variation in among their sequences. This variation is introduced by dynamic objects - pedestrians, cars, bicycles, trucks, vans. Due to the presence of multiple dynamic objects, correspondence pairs for CARLA-64 and ARD-16 also have significant variation.

\subsection{Experimental Setup}

We give the details of the experimental setup for static reconstruction using \dslr{}. All models are trained on an NVIDIA RTX-3090Ti GPU. \dslr{} is trained for 1500 epochs, using a batch size of 64 with RMS optimizer. We use a DCGAN-based generator and a MLP-based discriminator. The learning rate is initialized to 1e-4 for the network. \dslruda{} is initialized with CARLA-64 pre-trained weights and is trained for 600 epochs using similar model parameters as above.

\dslr{} acts as a preprocessing step on LiDAR scans before they used for downstream tasks. We use Google Cartographer (\citet{hess2016real}), a well known LiDAR-based SLAM algorithm to test the LiDAR sequences. The experiments are run on Intel(R) Core(TM) i5-271
4590 CPU @ 3.30GHz processor with 16 GB of RAM using ROS Noetic Distro.  The inference time for a single LiDAR scan for our model is 5ms which allows it to work in real time.

\subsection{Baselines}

\subsubsection{Baselines for Static LiDAR Reconstruction }
\label{baseline-recon}

We choose models that have shown impressive performance on the problem of static scene reconstruction for LiDAR, images without using segmentation labels and on the problem of shape completion for point clouds. It is accepted in literature (\citet{bescos2019empty}) that learning based methods outperform non-learning based methods. We do not use LiDARGen, NeRF-LiDAR, Ultra-LiDAR as baselines due to reasons explained in Section \ref{related}.
The following methods show reasonable to strong performance on LiDAR for our task. Thus, we use them as our baselines: 
1). \CacciaAE{} (C-AE) 2). \CacciaGAN{} (C-GAN) \cite {caccia2018deep} 3). \Coarse{}  (C-NET) \cite{wang2022coarse} 4). \siv{} (S-INV) \cite{zhang2021unsupervised} 5). \ds{} \cite{kumar2021dynamic}. For details on these baselines, please refer to Appendix - Section 1.

\subsubsection{Baselines for Navigation Evaluation}
\label{baseline-slam-eval}

We compare the performance of \dslr{} on navigation against segmentation-free as well as segmentation-based baselines.

For segmentation-free settings, we choose the model from Section \ref{baseline-recon} which generates the best static reconstruction, as our baseline. We use the generated static reconstructions that replace the dynamic objects in LiDAR.
 
\dslr{} operate in segmentation information free settings. For the purpose of having a better insight, we also attempt a comparison of \dslr{}'s with settings where segmentation information is available. We compare \dslr{} against a strong segmentation-based method - KITTI-Seg in Table \ref{tab:kitti-seg-all}.  KITTI-Seg utilizes groundtruth semantic segmentation information to remove the dynamic objects from the KITTI sequences. We show these results on KITTI dataset because of the availability of KITTI semantic segmentation information.

\subsection{Evaluation Metrics}
\label{metrics}
\subsubsection{Static Reconstruction Evaluation}
\label{metrics}
We use Chamfer Distance (\citet{fan2017point}) and Earth Mover's Distance (\citet{emd-cd}) to evaluate the closeness of the reconstructed static scans to the corresponding original static LiDAR scans for CARLA-64 and ARD-16. For KITTI, we adopt the LQI metric. We provide the details of these baselines below -

\textbf{Earth Mover's Distance}: The Earth Mover's Distance between two given point clouds $PC_1$ and $PC_2$ calculates the least amount of work requires to transform one to the another.
    Formally, it can be written as
    \begin{equation}
    \begin{split}
        =\min _{\phi: PC_{1} \longrightarrow PC_{2}} \sum_{x \in PC_{1}}\|x-\phi(x)\|_{2}
            \end{split}
        \end{equation}
        Here, $\phi$ represents a bijective mapping between $PC_{1}$ and $PC_{2}$. 
    To calculate EMD between the two point clouds  we require a mapping that minimizes the above term. Calculating EMD is computationally expensive as it requires both point clouds to have an equal number of points.

\textbf{Chamfer's Distance}: Given 2 set of points clouds $PC_1$ and $PC_2$, Chamfer's Distance can be expressed as 
        \begin{equation}
            \begin{split}
          =\sum_{x1 \in PC_{1}} \min _{x2 \in PC_{2}}\|x1-x2\|_{2}^{2}+\sum_{x2 \in PC_{2}} \min _{x1 \in PC_{1}}\|x1-x2\|_{2}^{2}
          \end{split}
        \end{equation}
    The approach involves identifying the closest point in $PC_2$ for each point in $PC_1$ and vice versa, and adding the obtained distances. Unlike EMD,  it allows a single point in $PC_1$ to map to multiple points in $PC_1$, and vice versa. As a result, the two point clouds may have different numbers of points.

\textbf{LiDAR Quality Index}:
LiDAR Quality Index (LQI) is a metric that measures the level of noise in a LiDAR scan. It is based on  work done on No-Reference Image Quality Assessment where the visual quality of an image is evaluated without any reference image or knowledge of the distortions present in the image (\citet{kang2014convolutional}).  It is based on the assumption that LiDAR data follows a certain distribution and dynamic objects are patterns of external noise in that distribution. A lower LQI score indicates that the scan is closer to a static scan and has no noise. This metric is particularly useful in measuring the quality of reconstructed static KITTI scans when groundtruth static KITTI scans are not available. LQI has a positive correlation with Chamfer's distance.
 (\citet{kumar2021dynamic}) state that scans taken in a static environment have lower values for both Chamfer's Distance and LQI compared to scans taken in dynamic environments. It suggests that scans with lower LQI scores have less dynamic movement and are closer to scans taken in a static environment. For more details on LQI, please refer to Appendix - Section 2.

\subsubsection{Navigation Evaluation Metrics}
We compare the pose-estimates/trajectories generated by Google Cartographer using the reconstructed static LiDAR structures, with the ground truth pose estimates using 2 widely used metrics - Absolute Trajectory Error (ATE)  and Relative Pose Error (RPE).

\textbf{Absolute Trajectory Error}: 
ATE is used for measuring the overall global consistency between two trajectories. It focuses on measuring the difference between the translation components of the two trajectories. This is accomplished in two steps - \textbf{(1)} Aligning the coordinate frames for both trajectories \cite{horn1987closed} and \textbf{(2)} Comparing the absolute distances between the translational components of the individual pose estimates of both the trajectories.

\textbf{Relative Pose Error}: 
RPE is used to determine the amount of drift present in a trajectory \textit{w.r.t.} a ground truth trajectory, locally. It is robust against the globally accumulated error. It measure the discrepancy between the estimated and ground truth trajectories over a specified time interval ($\delta$).  By using $n$ poses, we compute $m$ individual RPE values (\textit{m = n }- $\delta$), which are aggregated to calculate the overall error for the trajectory. RPE takes into account both the translational and rotational components of the trajectory.

For further information on the metrics, please refer to the Sturm et. al.\cite{sturm2012benchmark}.
\section{Results}
\subsection{
Static Translation using \dslr{}}
\label{sec:chamfer-nu}
CARLA-64 and ARD-16 have correspondence pairs available.
Hence, we report Chamfer's Distance and Earth Mover's Distance for them across the 5 baselines. (Table \ref{tab:carlaresults}). \dslr{}
performs better at both metrics against all the baselines.
% We also show ablation studies w.r.t. the components of our models over the best-performing  baseline in Table \ref{tab:ablation}. 
The key advantage of \dslr{} is that it is able to accurately differentiate between static and dynamic objects as well as between the corresponding occlusion patterns. It omits dynamic structures/occlusions and preserves static structures/occlusions (Video Demo at \cite{url1}). This effect is due to the explicit attention generated by the couple discriminator on dynamic objects and due to the strategy of contrasting LiDAR triplets using carefully chosen hard negatives. 
\setlength{\tabcolsep}{1.5pt}

 \begin{table*}[htbp]
\scriptsize
    \centering
    \begin{tabular}{|c|c|c|c|c|c|c|}

    \hline
     Run & C-Net         & S-INV        & C-AE         & C-GAN       &  Best-Baseline       & Ours        
\\ \hline
\multicolumn{7}{|c|}{\textbf{CARLA-64} (CD/EMD)} \\ \hline

 8 & 18.08/506        & 11.91/981  & 1.14/291.12  & 1.58/243.52 &  0.88/259.23 & \textbf{0.37/235.90} \\ \cline{1-7} 
 9 & 14.93/432     & 9.11/881   & 1.39/194.57  & 1.48/198.36 &  1.33/185.95 & \textbf{0.33/156.14}  \\ \cline{1-7} 
 10 & 17.43/692.32 & 10.23/887  & 6.06/483.59  & 5.11/463.12 &  1.57/231.37 & \textbf{1.19/226.81} \\ \cline{1-7} 
 11 & 6.20/608     & 8.56/981   & 1.92/324.26  & 1.59/292.21 &  1.24/239.97 & \textbf{0.95/207.79} \\ \cline{1-7} 
 12   & 17.91/597    & 9.32/901   & 3.89/356.20  & \textbf{3.54}/271.63 &  8.44/206.67 & 5.36/\textbf{186.35} \\ \cline{1-7} 
 13  &17.11/ 499.3  & 12.53/965  & 2.15/388.81  & 1.71/341.26 &  1.04/243.74 & \textbf{0.6/202.30}  \\ \cline{1-7} 
 14  &  16.18/501    & 14.12/1004 & 11.01/181.61 & 8.92/185.71 & \textbf{2.57}/228.34 & 2.63/\textbf{210.48}  \\ \hline
\multicolumn{7}{|c|}{\textbf{ARD-16} (CD/EMD)} \\ \hline

 1& 3.98/502     & 7.49/698  & 0.95/201  & 10.65/195 &  0.37/194 & \textbf{0.36/128} \\ \hline
\multicolumn{7}{|c|}{\textbf{KITTI} (LQI)} \\ \hline
% \MULTIROW{1}{*}{CARLA-64} 
 & 18.08/506        & 11.91/981  & 1.14/291.12  & 1.58/243.52 &  0.88/259.23 & \textbf{0.37/235.90} \\ \hline

    \end{tabular}
    \caption{Comparison of our model on static reconstructions with
5 baselines on 3 datasets. For all metrics lower is better. A video demo of the reconstructions is available at \cite{url1}.  }
    \label{tab:carlaresults}
    \end{table*}

We observe that our model works reasonably well even on
challenging sparse dataset - ARD-16. This is also highlighted by the fact that
\dslr{} is capable of generating precise dynamic object segmentation, unlike the best baseline (Figure \ref{fig:ati-moving-movable}) for ARD-16. \dslr{} overcomes the shortcoming of existing baselines by reconstructing only dynamic occlusion and retaining all other static structures. These static structures are vital for AVs in challenging scenarios like sparse datasets (ARD-16) on industrial floors.

For KITTI, we use LQI as a metric due to the unavailability of ground truth static scans. Our method improves to a large extent on the available baselines. The improvement in static translation translates to superior downstream  performance in segmentation-devoid settings (Table \ref{tab:slamresults}). 
% For a video demo of the static reconstruction results, click  \href{https://drive.google.com/file/d/1YDX4PBMQJMbuelSR_Ox7XjzXuOWu6bu_/view?usp=sharing}{here}.

\subsection{Segmentation of Dynamic Objects}
\label{sec:segm_mova_movi}

\begin{figure}[htbp]
  \centering
\includegraphics[width=0.95\linewidth, height=0.45\linewidth]{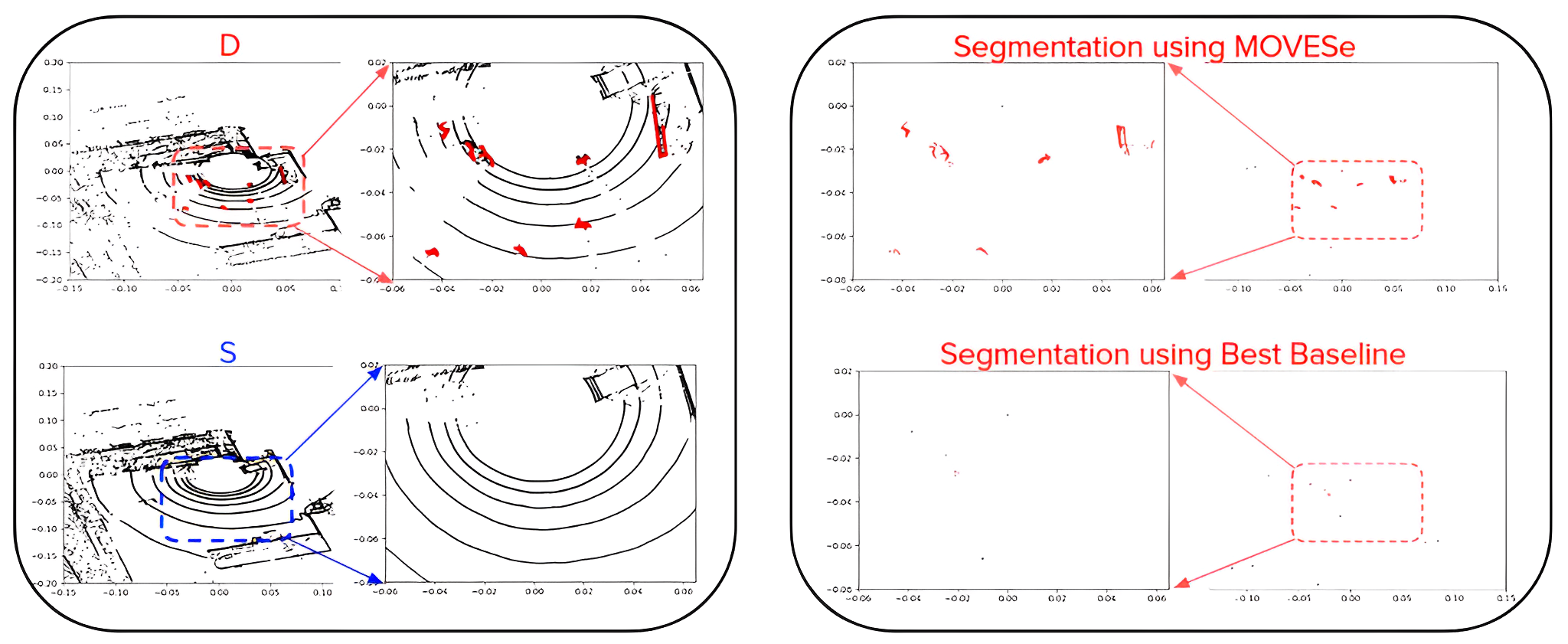}  
  \caption{Dynamic Object Segmentation comparison on ARD-16. \textbf{Left}: A corresponding LiDAR pair. Top is dynamic (\textcolor{red}{D}) and bottom is static (\textcolor{blue}{S}) scan. Note the presence of dynamic objects (\textcolor{red}{in red}) in the top dynamic scan (\textcolor{red}{D}). \textbf{Right}: Given dyanmic scan, \textcolor{red}{D} as input, segmentation output generated by our model is on top. Segmentation output by the best baseline is at bottom. \dslr{} precisely segments dynamic objects while the baseline is unable to capture them, which can be visualized in the Video demo \cite{url2}.} 

\label{fig:ati-moving-movable}
\end{figure}

\begin{figure}[t]
    \centering
    \includegraphics[width=6in]{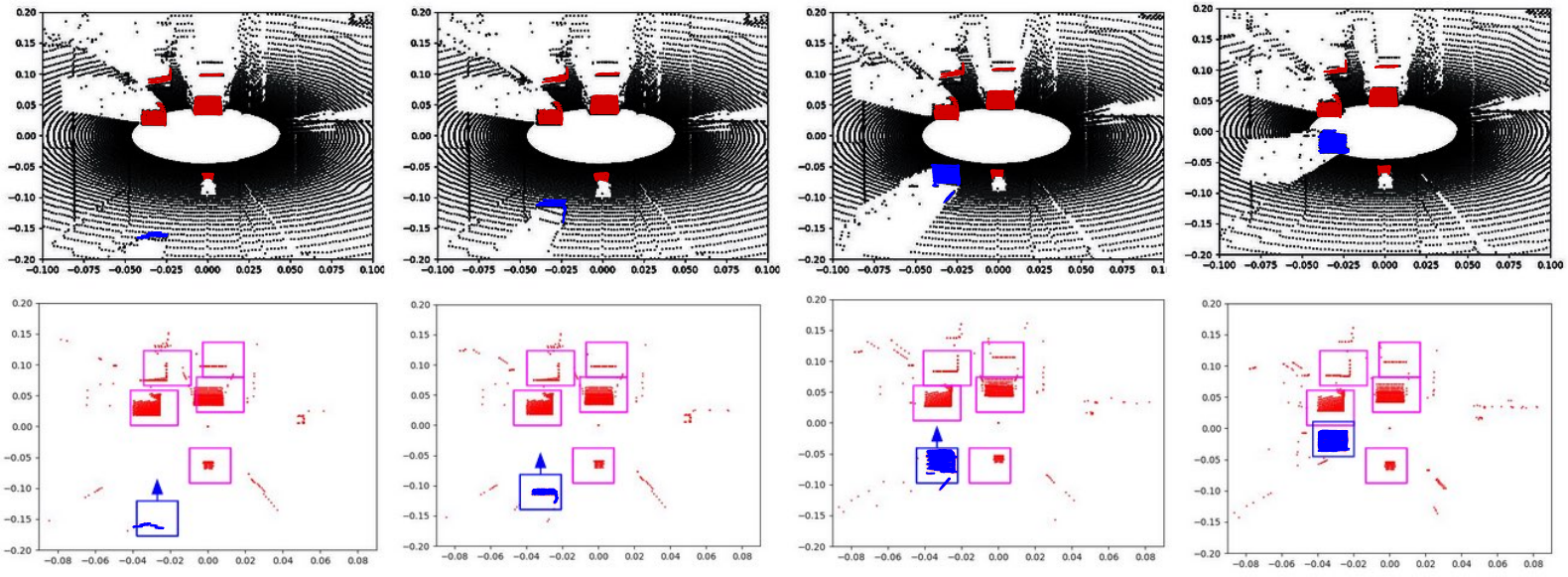}
    \caption{(Please view from Left to Right) \textbf{Row 1}: Top view of a contiguous  series of 4 dynamic scans at time intervals \textit{t, t+3, t+6, t+9} in a LiDAR sequence. Red denotes movable objects while blue denotes moving objects. 
    \textbf{Row 2}: When dynamic scan in row 1 is given as input to \dslr{}, it generates segmentation of movable objects (red) and moving objects (blue). Note that the moving object (\textcolor{blue}{in blue}) changes position across the scans from left to right while the movable objects (\textcolor{red}{in red}) are currently stationary. These movable objects, although stationary here, can be seen as moving in the video demo \cite{url3}.} 

    \label{fig:movable-second}
\end{figure}

We qualitatively show the segmentation of moving as well as movable objects in LiDAR scans (Figure \ref{fig:movable-first}, \ref{fig:ati-moving-movable} and \ref{fig:movable-second}) using \dslr{}. We achieve this by subtracting the original dynamic input from the generated static reconstruction (consisting of only static structures). The subtraction yields precise details of the non-stationary objects. These non-stationary objects can be
easily split into movable and movable by temporally observing the relative change in their position w.r.t the vehicle. The key advantage of \dslr{} is that unlike existing baselines, it accurately differentiates between dynamic and static obstacles and their patterns and removes only dynamic obstacles via static structure reconstruction. This results in accurate dynamic object segmentation (Figure \ref{fig:movable-second} and Video Demo at \cite{url3}). 
Moreover, challenging sparse datasets like ARD-16 have few points falling on dynamic objects. We show that the best baseline model fails to identify the dynamic objects for ARD-16. However, we are successful in segmenting out dynamic objects in such sparse datasets (Figure \ref{fig:ati-moving-movable} and Video Demo at \cite{url2}).

\subsection{Application of \dslr{} for navigation in dynamic environments}
In this section, we demonstrate the benefit of reconstructed static structures on navigation. Our model acts as a preprocessing step on LiDAR scans before they are consumed by the  navigation algorithm.

For KITTI, we use the KITTI Odometry dataset sequences from 00-10. For ARD-16, we use the single run released, which is longer and combines all sequences and scenes available in previous testing runs.  For CARLA-64, we use the 4 available dynamic LiDAR sequences.

We first compare \dslr{} against the best baselines in in the segmentation-free settings using all the three datasets in Table \ref{tab:slamresults}.

We also compare \dslruda{} with a segmentation-label assisted baseline - KITTI-Seg. KITTI-Seg utilizes semantic segmentation information to remove dynamic objects before using the LiDAR scans for navigation. We show these results on the KITTI dataset because of the availability of semantic segmentation information. The comparison of KITTI-Seg against \dslruda{} is available in Table \ref{tab:kitti-seg-all}.

\label{sec:slamindyenv}

\setlength{\tabcolsep}{1.5pt}
\begin{table}[t]
\scriptsize
\centering
\begin{tabular}{|ccccccc|}
\hline
\multicolumn{1}{|c|}{}           & \multicolumn{3}{c|}{\textbf{Best Baseline}}                                                                                            & \multicolumn{3}{c|}{\dslr{}}                                                                              \\ \hline
\multicolumn{1}{|c|}{Run}        & \multicolumn{1}{c|}{ATE}         & \multicolumn{2}{c|}{RPE}                                                          & \multicolumn{1}{c|}{ATE}            & \multicolumn{2}{c|}{RPE}                                          \\ \hline
\multicolumn{1}{|c|}{}           & \multicolumn{1}{c|}{(m)}         & \multicolumn{1}{c|}{Trans (m)}        & \multicolumn{1}{c|}{Rot ($\theta$)}         & \multicolumn{1}{c|}{(m)}            & \multicolumn{1}{c|}{Trans (m)}      & Rot ($\theta$)                \\ \hline
\multicolumn{1}{|c|}{}           & \multicolumn{6}{c|}{\textbf{CARLA-64}}                                                        \\ \hline
\multicolumn{1}{|c|}{0}          & \multicolumn{1}{c|}{5.67}        & \multicolumn{1}{c|}{0.102}            & \multicolumn{1}{c|}{0.413}                & \multicolumn{1}{c|}{\textbf{1.74}}  & \multicolumn{1}{c|}{\textbf{0.075}} & \textbf{0.412}              \\ \hline
\multicolumn{1}{|c|}{1}          & \multicolumn{1}{c|}{2.71}        & \multicolumn{1}{c|}{\textbf{0.050}}   & \multicolumn{1}{c|}{\textbf{0.390}}       & \multicolumn{1}{c|}{\textbf{1.10}}  & \multicolumn{1}{c|}{0.060}          & \textbf{0.390}              \\ \hline
\multicolumn{1}{|c|}{2}          & \multicolumn{1}{c|}{1.34}        & \multicolumn{1}{c|}{\textbf{0.077}}   & \multicolumn{1}{c|}{\textbf{0.526}}       & \multicolumn{1}{c|}{\textbf{1.11}}  & \multicolumn{1}{c|}{0.078}          & \textbf{0.520}              \\ \hline
\multicolumn{1}{|c|}{3}          & \multicolumn{1}{c|}{6.80}        & \multicolumn{1}{c|}{\textbf{0.136}}   & \multicolumn{1}{c|}{\textbf{0.375}}       & \multicolumn{1}{c|}{\textbf{3.83}}  & \multicolumn{1}{c|}{0.212}          & 0.379                       \\ \hline
\multicolumn{1}{|c|}{}           & \multicolumn{6}{c|}{\textbf{ARD-16}}                      \\ \hline
\multicolumn{1}{|c|}{0}          & \multicolumn{1}{c|}{2.06}        & \multicolumn{1}{c|}{\textbf{0.067}}   & \multicolumn{1}{c|}{1.550}                & \multicolumn{1}{c|}{\textbf{1.97}}  & \multicolumn{1}{c|}{\textbf{0.067}} & \textbf{1.520}                       \\ \hline
\multicolumn{1}{|c|}{}           & \multicolumn{6}{c|}{\textbf{KITTI}}                      \\ \hline

\multicolumn{1}{|c|}{0}          & \multicolumn{1}{c|}{5.22}        & \multicolumn{1}{c|}{0.025}            & \multicolumn{1}{c|}{0.005}                & \multicolumn{1}{c|}{\textbf{2.64}}  & \multicolumn{1}{c|}{\textbf{0.015}} & \textbf{0.003}              \\ \hline
\multicolumn{1}{|c|}{1}          & \multicolumn{1}{c|}{24.64}       & \multicolumn{1}{c|}{0.140}            & \multicolumn{1}{c|}{0.007}                & \multicolumn{1}{c|}{\textbf{4.91}}  & \multicolumn{1}{c|}{\textbf{0.080}} & \textbf{0.003}              \\ \hline
\multicolumn{1}{|c|}{2}          & \multicolumn{1}{c|}{8.98}        & \multicolumn{1}{c|}{0.032}            & \multicolumn{1}{c|}{0.003}                & \multicolumn{1}{c|}{\textbf{5.84}}  & \multicolumn{1}{c|}{\textbf{0.026}} & \textbf{0.002}              \\ \hline
\multicolumn{1}{|c|}{4}          & \multicolumn{1}{c|}{1.85}        & \multicolumn{1}{c|}{0.158}            & \multicolumn{1}{c|}{0.132}                & \multicolumn{1}{c|}{\textbf{0.77}}  & \multicolumn{1}{c|}{\textbf{0.108}} & \textbf{0.124}              \\ \hline
\multicolumn{1}{|c|}{5}          & \multicolumn{1}{c|}{2.05}        & \multicolumn{1}{c|}{0.026}            & \multicolumn{1}{c|}{\textbf{0.004}}       & \multicolumn{1}{c|}{\textbf{1.30}}  & \multicolumn{1}{c|}{\textbf{0.015}} & 0.005                       \\ \hline
\multicolumn{1}{|c|}{6}          & \multicolumn{1}{c|}{4.46}        & \multicolumn{1}{c|}{0.263}            & \multicolumn{1}{c|}{0.025}       & \multicolumn{1}{c|}{\textbf{1.82}}  & \multicolumn{1}{c|}{\textbf{0.060}} & \textbf{0.013}                       \\ \hline
\multicolumn{1}{|c|}{7}          & \multicolumn{1}{c|}{5.23}        & \multicolumn{1}{c|}{0.026}            & \multicolumn{1}{c|}{0.006}                & \multicolumn{1}{c|}{\textbf{2.64}}  & \multicolumn{1}{c|}{\textbf{0.016}} & \textbf{0.003}              \\ \hline
\multicolumn{1}{|c|}{8}          & \multicolumn{1}{c|}{6.70}        & \multicolumn{1}{c|}{0.031}            & \multicolumn{1}{c|}{\textbf{0.006}}       & \multicolumn{1}{c|}{\textbf{2.90}}  & \multicolumn{1}{c|}{\textbf{0.018}} & \textbf{0.006}              \\ \hline
\multicolumn{1}{|c|}{9}          & \multicolumn{1}{c|}{27.96}        & \multicolumn{1}{c|}{0.350}            & \multicolumn{1}{c|}{0.114}                & \multicolumn{1}{c|}{\textbf{4.50}}  & \multicolumn{1}{c|}{\textbf{0.055}} & \textbf{0.043}              \\ \hline
\multicolumn{1}{|c|}{10}         & \multicolumn{1}{c|}{2.45}        & \multicolumn{1}{c|}{0.034}            & \multicolumn{1}{c|}{0.010}                & \multicolumn{1}{c|}{\textbf{0.70}}  & \multicolumn{1}{c|}{\textbf{0.024}} & \textbf{0.008}              \\ \hline
\end{tabular}
\caption{Comparison of SLAM Results on 4 CARLA-64, 1-ARD and 10 KITTI Sequences without segmentation. For all metrics, lower is better.}
\label{tab:slamresults}
\end{table}

 \subsubsection{Analysis of navigation Results}

\textbf{Segmentation-devoid scenarios} - In Table \ref{tab:slamresults}, we demonstrate that in segmentation devoid settings, \dslr{} is able to improve navigation performance when compared against the best baseline over all the three datasets. This results deomonstrate the importance of the accurate static structures generated by \dslr{}. For KITTI we do not have corresponding static-dynamic LiDAR scans available. Inspite of this shortcoming, \dslruda{} is able to achieve superior results. This demonstrates that our approach can be successfully applied to datasets even in the absence of  correspondence information in label-devoid settings.

We notice that the improvement in navigation error for ARD-16 is marginal compared to the other datasets (Table \ref{tab:slamresults}). This is expected because ARD-16 is a sparse dataset with low density of points falling on the dynamic objects. The \textit{new static points} in the reconstructed static scan that replace the dynamic points are fewer in number.  The benefit on navigation is directly proportional
to the number of new static points introduced in the generated scan.
Therefore, the improvement on ARD-16 is expected to be less compared to the dense datasets. 
\begin{figure}[htbp]
% \centering
\begin{tabular}{|c|c|c|}
\hline
\subf{\includegraphics[width=0.33\linewidth]{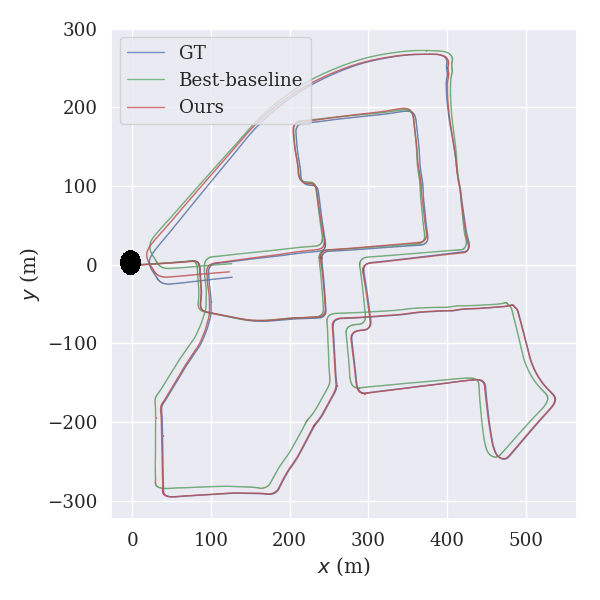}}
     {Sequene 00}
&
\subf{\includegraphics[width=0.33\linewidth]{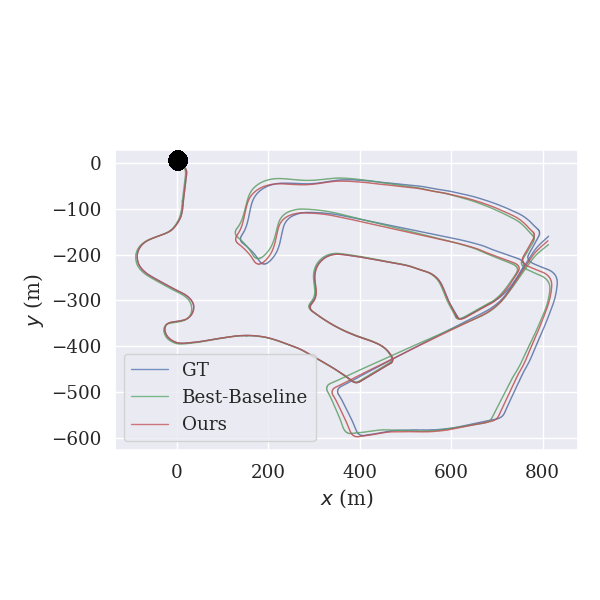}}
     {Sequene 02}

&
\subf{\includegraphics[width=0.33\linewidth]{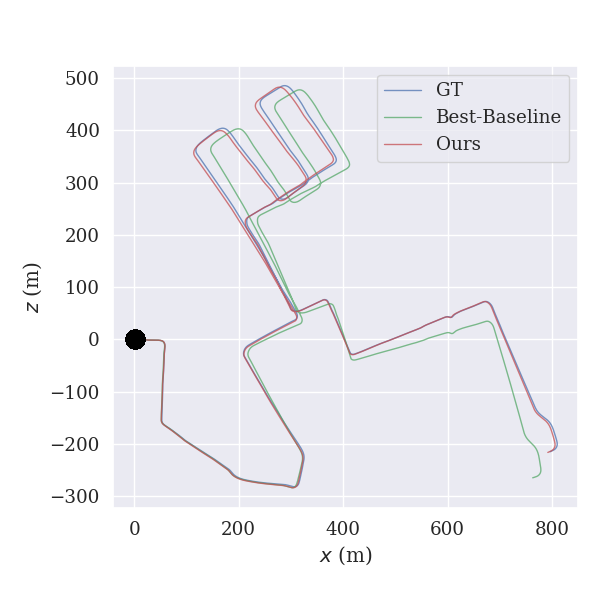}}
     {Sequene 08}
\\

\hline
\subf{\includegraphics[width=0.33\linewidth]{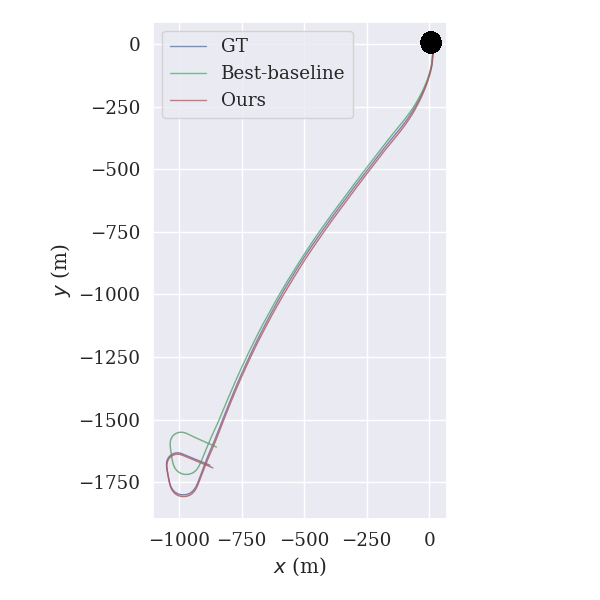}}
     {Sequene 01}
&

\subf{\includegraphics[width=0.33\linewidth]{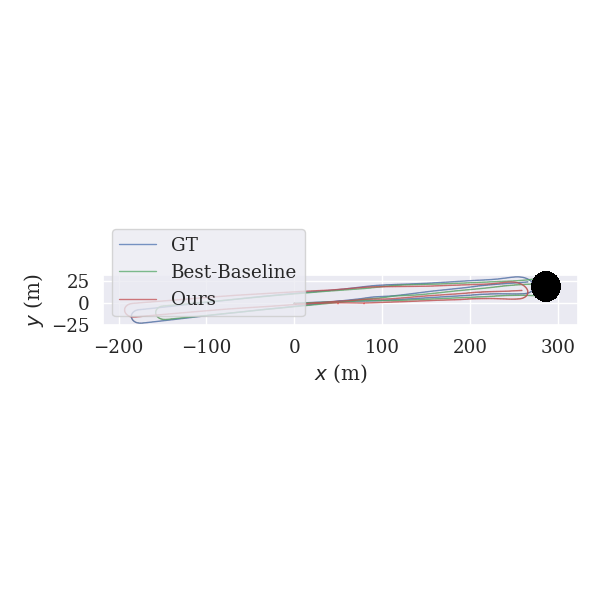}}
     {Sequene 06}
&
\subf{\includegraphics[width=0.33\linewidth]{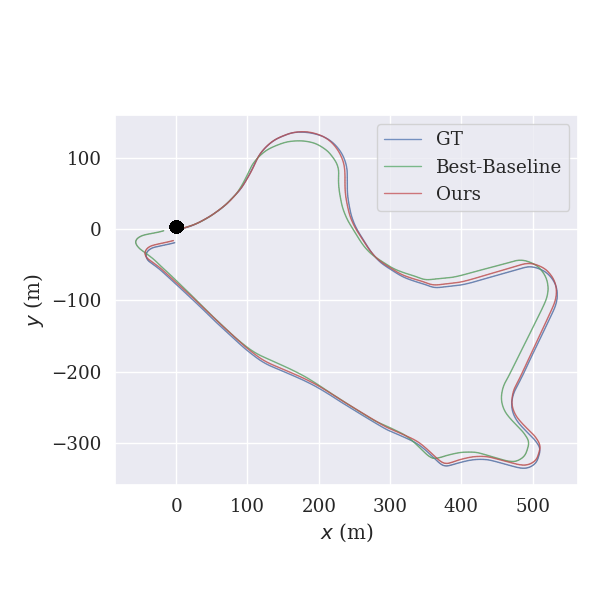}}
     {Sequene 09}

\end{tabular}

\caption{Comparison of the estimated trajectory using \dslr{} with the best baseline for 6 sequences. \textbf{Row 1}- sequences with multiple loop closures. \textbf{Row 2} - Sequences with minimal (0 or 1)  loop closures.}
\label{fig:kitti-slam-trajectory}
\end{figure}

\begin{table}[]
\scriptsize
\centering

\begin{tabular}{|c|c|c|c|}
\hline
\multirow{2}{*}{Sequence} & \multirow{2}{*}{Loop Clousres} & \multirow{2}{*}{\begin{tabular}[c]{@{}c@{}}Dynamism Level\\ (Number of dynamic objects)\end{tabular}} & \multirow{2}{*}{\begin{tabular}[c]{@{}c@{}}Lenght\\ (Number of LiDAR scans)\end{tabular}}  \\
                          &                                &             &                                                                                          \\ \hline
0                         & \textgreater{}=2               & Medium(21)  &        4390                                                                             \\ \hline
1                         & 1                              & High(53)    &        1173                                                                                \\ \hline
2                         & \textgreater{}=2               & Low(9)      &        4485                                                                               \\ \hline
4                         & 0                              & High(30)    &        284                                                                              \\ \hline
5                         & \textbf{\textgreater{}=2}               & \textbf{High(40)}    &        \textbf{2750}                                                                             \\ \hline
6                         & 1                              & Low(10)     &        1091                                                                            \\ \hline
7                         & 1                              & Medium(18)  &        1111                                                                           \\ \hline
8                         & \textbf{\textgreater{}=2}               & \textbf{High(49)}    &        \textbf{5149}                                                                          \\ \hline
9                         & 1                              & Low(13)     &        1599                                                                         \\ \hline
10                        & 0                              & Low(10)     &        1215                                                                        \\ \hline
\end{tabular}
\caption{Variation observed in the KITTI sequences w.r.t. number of loop closures and level of dynamism - associated with number of dynamic objects and length of sequence.}
\label{tab:kitti-variation}
\end{table}

We compare the estimated trajectories of \dslr{} with the best baseline on the urban KITTI Odometry sequence.
The KITTI Odometry sequences are diverse on several parameters - loop closures, length, dynamic crowded urban sequences to sparsely dynamic highway sequences. We show the variations in Table \ref{tab:kitti-variation}. Sequences 0, 2, 5, 8 have multiple loops closure while the other sequences have less than 2 loop closures. Sequences 1, 4, 5, 8 have high dynamism compared to the other sequences.

We compare the estimated trajectories of \dslr{} with the best baseline in Figure \ref{fig:kitti-slam-trajectory}. We observe a trend across the estimated  trajectories for the baseline - the navigation accumulates translation errors consistently as the sequence progresses despite loop closures. The top row in Figure \ref{fig:kitti-slam-trajectory} compares sequences with multiple loop closures. The baseline (indicated by green line) continues to accumulate translation error despite multiple loop closures as the run progresses from the start point (indicated by black dot). \dslr{} (indicated by red) is able to narrow down the error on these sequences. We observe a similar behaviour for sequences with minimal to no loop closures (row 2 of Figure \ref{fig:kitti-slam-trajectory}).

\begin{figure}

% \centering
\begin{tabular}{|c|c|c|}
\hline
\subf{\includegraphics[width=0.33\linewidth]{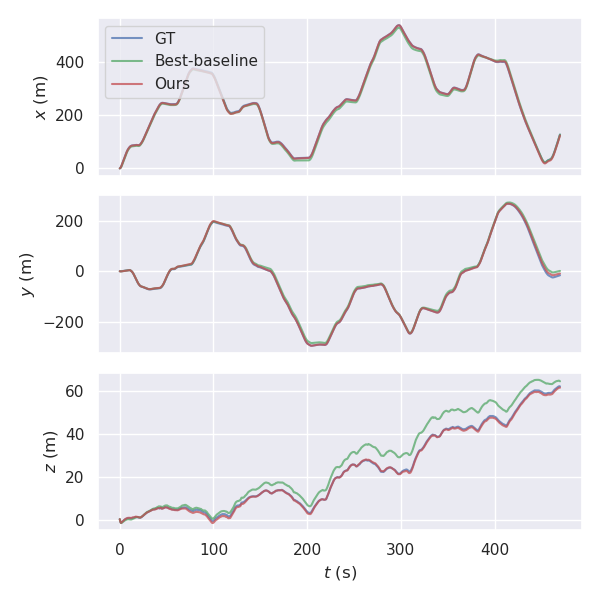}}
     {Sequene 00}
&
\subf{\includegraphics[width=0.33\linewidth]{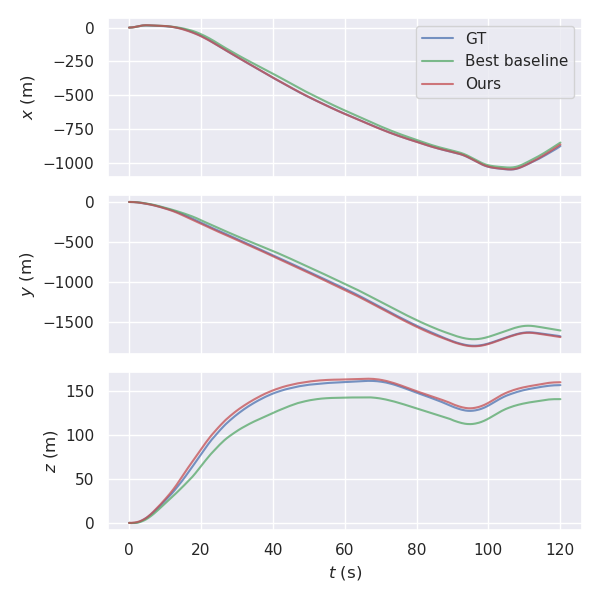}}
     {Sequene 01}

&
\subf{\includegraphics[width=0.33\linewidth]{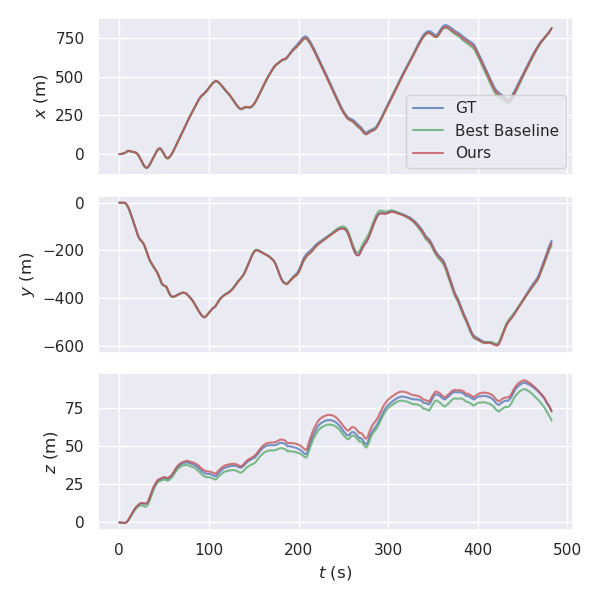}}
     {Sequene 02}
\\
\hline
\subf{\includegraphics[width=0.33\linewidth]{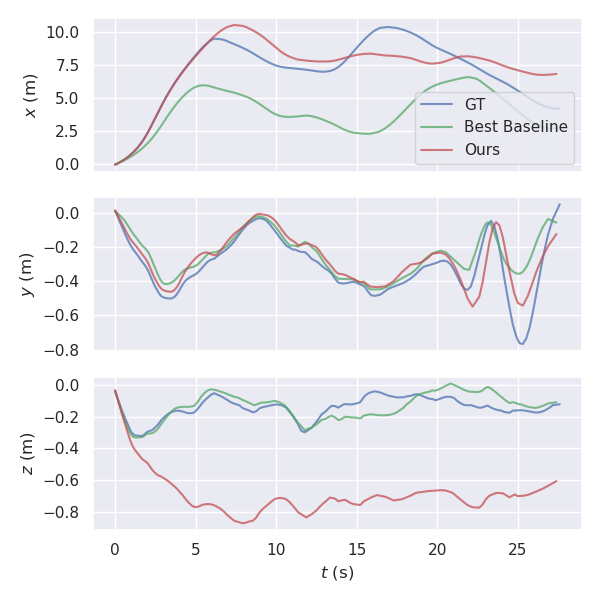}}
     {Sequene 04}
&
\subf{\includegraphics[width=0.33\linewidth]{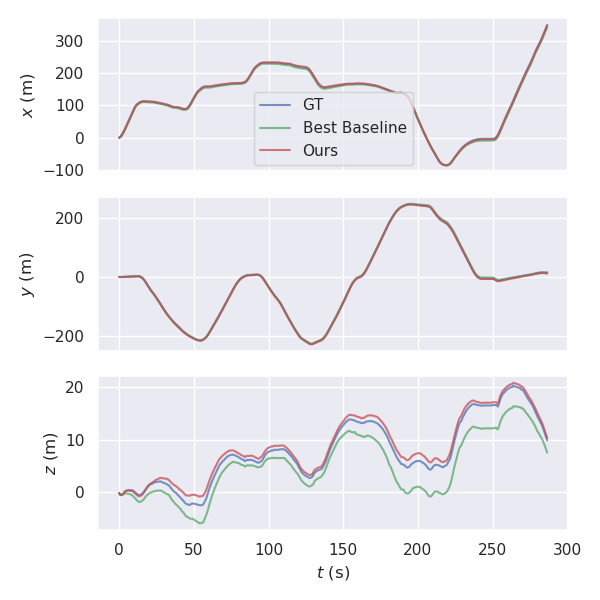}}
     {Sequene 05}

&
\subf{\includegraphics[width=0.33\linewidth]{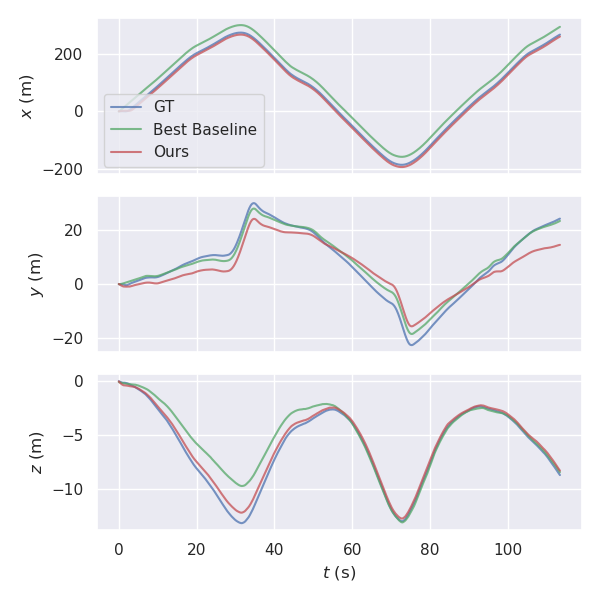}}
     {Sequene 06}
\\

\hline
\subf{\includegraphics[width=0.33\linewidth]{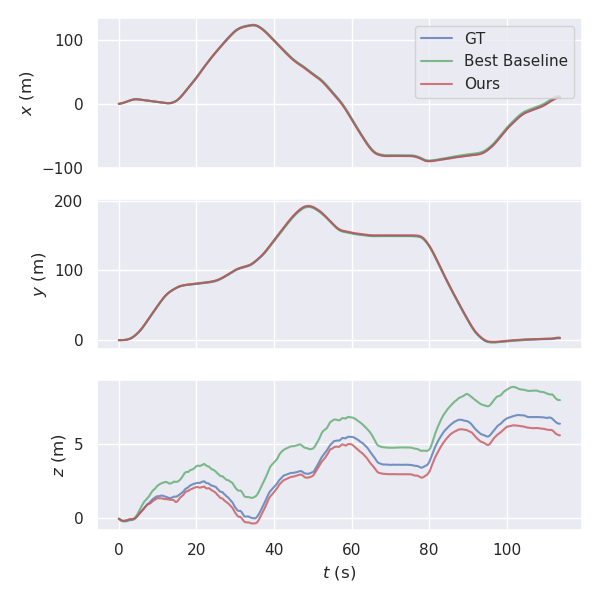}}
     {Sequene 07}
&
\subf{\includegraphics[width=0.33\linewidth]{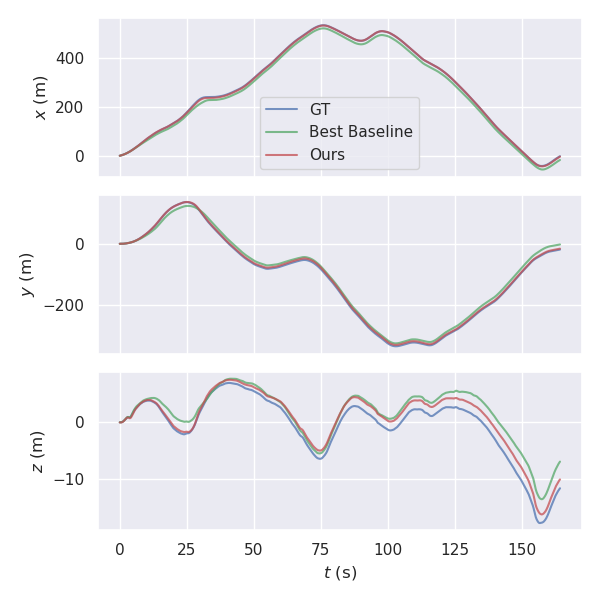}}
     {Sequene 09}

&
\subf{\includegraphics[width=0.33\linewidth]{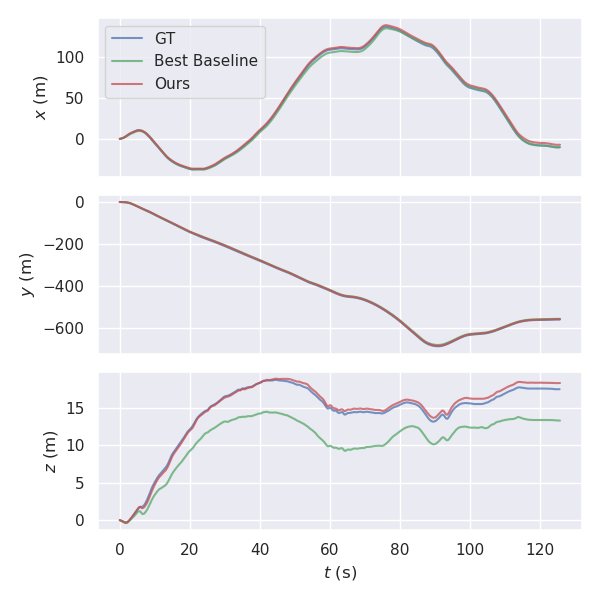}}
     {Sequene 10}
\\
\hline

\end{tabular}

\caption{ SLAM error along the x, y, z coordinates for KITTI sequences. Blue denotes GT, Red denotes Ours while Green Denotes Best baseline. We observe that the best baseline constantly has high errors along the z-dimension as compared to MOVES for most of the sequences. Refer to Table 2 for a quantitative analysis of the translation component of navigation error.}
\label{fig:kitti-slam-xyz-analysis}
\end{figure}

We investigate the reason behind the performance gain of \dslr{} using the KITTI dataset in Figure \ref{fig:kitti-slam-xyz-analysis}. We  plot the translation pose estimates of the best baseline and \dslr{} \textit{w.r.t.} the ground truth pose estimates for the KITTI sequences. Our analysis suggests that the baseline faces difficulty in accurately regressing the z-values for LiDAR scans during static reconstructions while leads to high pose error along the z-axis. Notably, it has significant errors along the z-axis, in addition to errors in the x and y axes. \dslr{} reduces the pose error along the z-axis and also reduces the error along the x and y axes for most sequences.

\begin{figure}

% \centering
\begin{tabular}{|c|c|c|}
\hline
\subf{\includegraphics[width=0.33\linewidth]{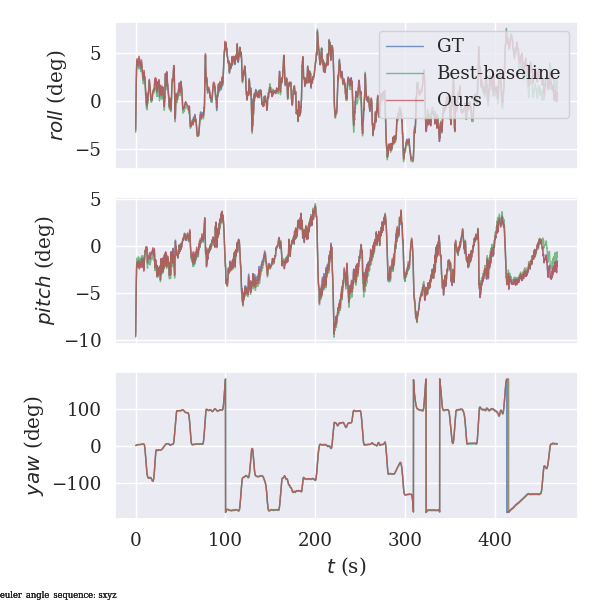}}
     {Sequene 00}
&
\subf{\includegraphics[width=0.33\linewidth]{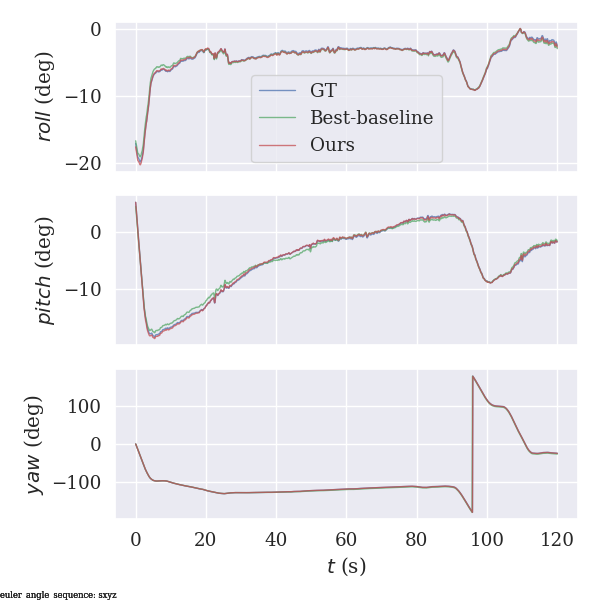}}
     {Sequene 01}

&
\subf{\includegraphics[width=0.33\linewidth]{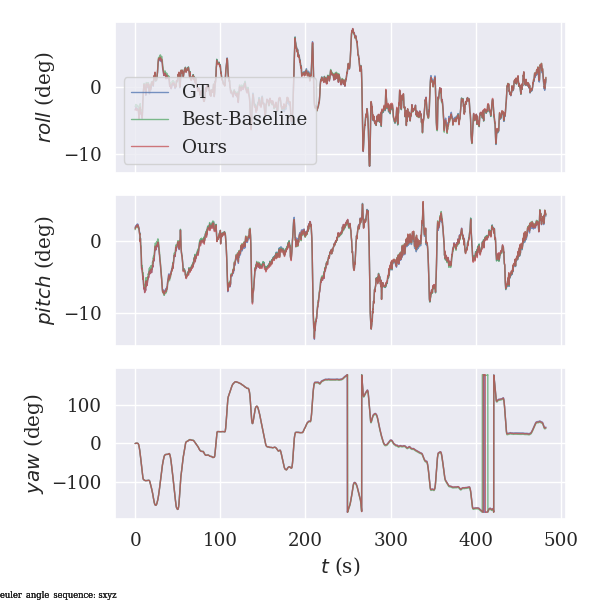}}
     {Sequene 02}
\\
\hline
\subf{\includegraphics[width=0.33\linewidth]{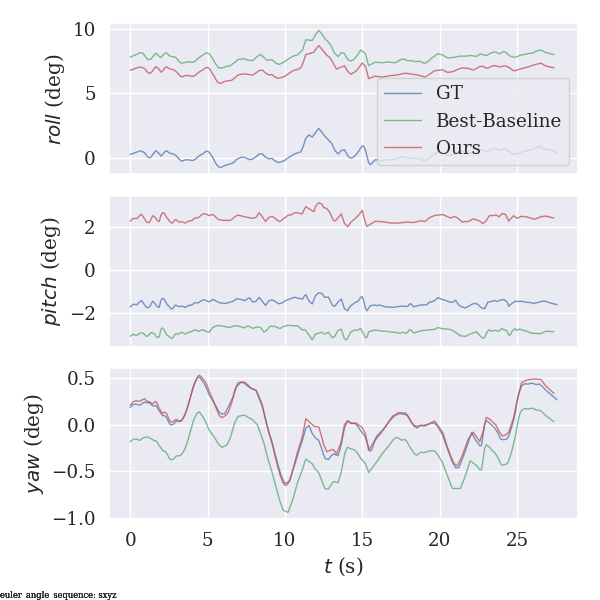}}
     {Sequene 04}
&
\subf{\includegraphics[width=0.33\linewidth]{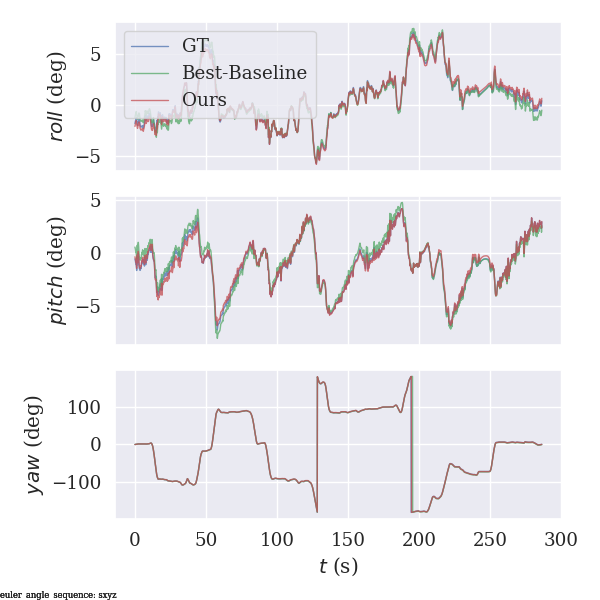}}
     {Sequene 05}

&
\subf{\includegraphics[width=0.33\linewidth]{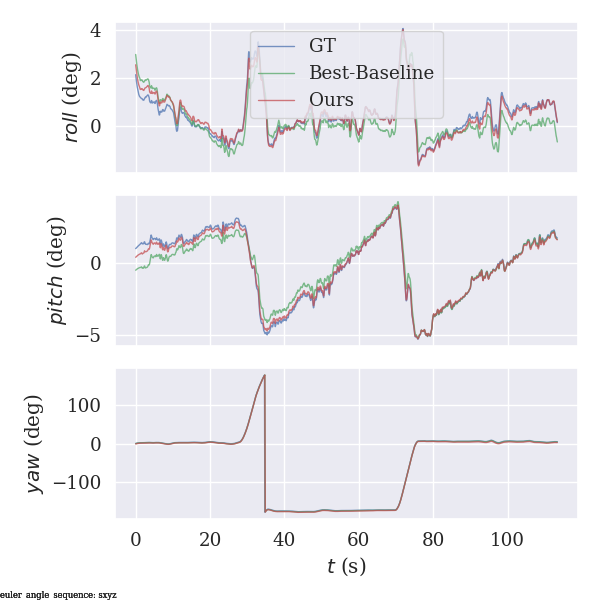}}
     {Sequene 06}
\\

\hline
\subf{\includegraphics[width=0.33\linewidth]{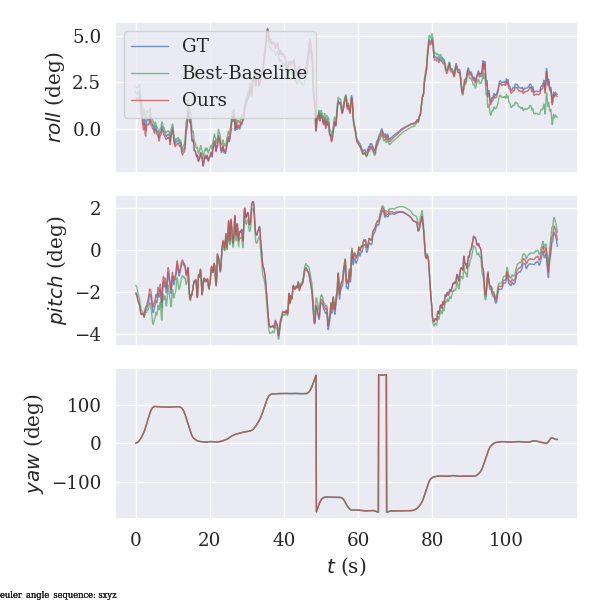}}
     {Sequene 07}
&
\subf{\includegraphics[width=0.33\linewidth]{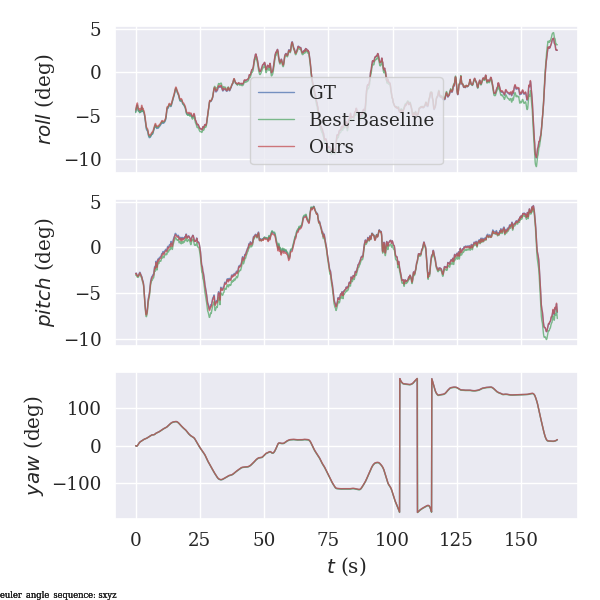}}
     {Sequene 09}

&
\subf{\includegraphics[width=0.33\linewidth]{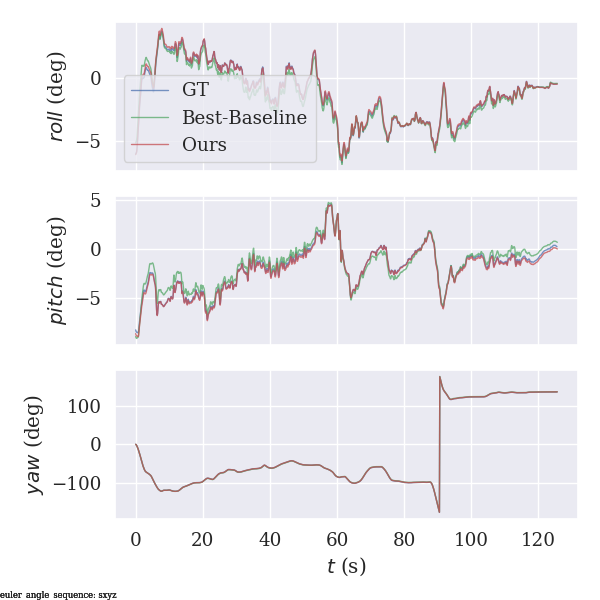}}
     {Sequene 10}
\\
\hline

\end{tabular}

\caption{ Navigation error along the rotational pose estimates (roll, pitch, yaw) for KITTI sequences. Blue denotes GT, Red denotes Ours while Green Denotes Best baseline. \dslr{} estimates the rotational poses closer to the ground truth consistently across the KITTI sequences. Refer to Table 2 for a quantitative analysis of rotational component of navigation error.}
\label{fig:kitti-slam-rpy-analysis}
\end{figure}

We also investigate the rotational pose estimates (roll, pitch, yaw) for the KITTI sequences and compare the performance of the baseline against \dslr{} in Figure \ref{fig:kitti-slam-rpy-analysis}. We observe that although the baseline works on par with \dslr{} for the yaw estimates, it performs sub-optimally for roll and pitch estimates across the KITTI sequences. \dslr{} consistently has lower Relative Pose error compared to the baseline (Table \ref{tab:slamresults} and Figure \ref{fig:kitti-slam-rpy-analysis}).

\textbf{Comparison of segmentation-assisted method (KITTI-Seg) against MOVES-MMD} - We compare the navigation performance of \dslruda{} (which works without segmentation assistance), against KITTI-Seg that utilizes segmentation information to remove dynamic objects, in Table \ref{tab:kitti-seg-all}. 

 \dslruda{} works without segmentation labels. As expected, it is not always superior compared to KITTI-Seg.
We surprisingly observer that for sequence 5, 8, \dslruda{} performs superior to KITTI-Seg on ATE without the availability of segmentation labels. These sequences have very \textit{high dynamism}, are relatively \textit{longer} and hence have \textit{multiple loop closures} unlike other sequences (Table \ref{tab:kitti-variation}). These are very practical situation that most AVs come across frequently in everyday scenarios. Our model excels in these settings. Our investigation suggest that sequences with high dynamism have fewer static structures. KITTI-Seg removes dynamic objects and occlusions. The LiDAR scans are left with the few static structures. In such cases, KITTI-Seg pose estimates are not robust to errors accumulated along the sequence, especially at turns where even consecutive static scans have relatively less overlap.  On the contrary, \dslruda{} not only removes dynamic objects and occlusion but also introduces multiple new static structures that replace the dynamic objects and corresponding occlusions. This leads to an overall increase in the amount of static content in LiDAR scans. These additions lead to robust pose estimation as opposed to \textit{highly dynamic settings} that have fewer static structures. Thus, the LiDAR estimated poses are highly resistant to drift based errors (that accumulate mostly at turns) due to robust scan matching using multiple static structures. This robustness is reinforced by the presence of several loop closures that ensure that the accumulated drift is minimized. Thus, longer KITTI sequences with high dynamism and multiple loop closures demonstrate better navigation performance using \dslruda{}, even in the absence of segmentation information.
\setlength{\tabcolsep}{0.8pt}
\begin{table}[H]
\scriptsize
\centering
\begin{tabular}{|c|c|c|c|c|}
\hline
\textbf{Run}                 & \textbf{Model}                        & \textbf{ATE}            & \multicolumn{2}{|c|}{{\textbf{RPE}}}   \\ \hline
                 &                         &       \textbf{Trans(m)}      &  \textbf{Trans(m)}      & \textbf{Rot($\theta$)}        \\ \hline
\multirow{2}{*}{0}  
& KITTI-Seg            &     \textbf{0.33}          & \textbf{0.008}          & 0.013          \\  
& \dslruda{} (No Seg)              & 2.64  & 0.015 & \textbf{0.003}  \\ 
\hline
\multirow{2}{*}{1}  
& KITTI-Seg            & \textbf{0.98}         & 0.024          & \textbf{0.002}          \\  
& \dslruda{} (No Seg)              &     4.91   &  \textbf{0.08}   &    0.003 \\\hline

\multirow{2}{*}{2} 
& KITTI-Seg            & \textbf{3.74}         & \textbf{0.008}          & 0.004           \\  
& \dslruda{} (No Seg)              & 5.84   & 0.026 & \textbf{0.002} \\\hline

\multirow{2}{*}{4}
& KITTI-Seg            &         \textbf{0.38} & \textbf{0.085}          & \textbf{0.070}          \\  
& \dslruda{} (No Seg)              & 0.77           & 0.108 & 0.124 \\\hline
\multirow{2}{*}{5}

& KITTI-Seg            &     1.85          & \textbf{0.005}          & \textbf{0.003}          \\  
& \dslruda{} (No Seg)              & \textbf{1.30} & 0.015 & 0.005 \\\hline
\multirow{2}{*}{6}  
& KITTI-Seg            &        \textbf{0.30}         &    \textbf{0.012}        &   \textbf{0.002}        \\  
& \dslruda{} (No Seg)              & 1.82           & 0.06  & 0.013 \\\hline
\multirow{2}{*}{7}  

& KITTI-Seg            &       \textbf{0.42}          & \textbf{0.005}          & 0.004          \\  
& \dslruda{} (No Seg)              & 2.64           & 0.016 & \textbf{0.003} \\\hline

\multirow{2}{*}{8}  

& KITTI-Seg            &       81.82          &   0.316          & 0.135          \\  
& \dslruda{} (No Seg)              & \textbf{2.64}           & \textbf{0.016} &         \textbf{0.003} \\\hline

\multirow{2}{*}{9} 
& KITTI-Seg            &       \textbf{1.35}         & \textbf{0.005}          & \textbf{0.003}          \\  
& \dslruda{} (No Seg)              & 4.50       & 0.055 &    0.043 \\\hline
\multirow{2}{*}{10}
& KITTI-Seg            &     \textbf{0.39}          & \textbf{0.009}           & \textbf{0.006}          \\  
& \dslruda{} (No Seg)              & 0.70   & 0.024 & 0.008 \\\hline
\end{tabular}
\caption{Detailed Comparison of MOVES (without segmentation) against segmentation baselines (KITTI-Seg) on KITTI sequences.}
\label{tab:kitti-seg-all}
\end{table}

\subsection{Ablation Studies}
\label{sec:ablation}

 \begin{table}[h]
 \scriptsize
    \centering
    \begin{tabular}{|c|c|c|c|}
    \hline
    Run & Best Baseline & CoD & CoD+CL \\ \hline
    & \multicolumn{3}{|c|}{{CD/EMD}}      \\ \hline

    8   & 0.88/259 & 0.52/\textbf{232}   & \textbf{0.38}/236  \\ \hline
    9   & 1.33/186  & 0.50/164  & \textbf{0.30}/\textbf{156}\\\hline

    10  & 1.57/231 & 1.26/\textbf{221}   &  \textbf{1.16}/227 \\ \hline
    11  & 1.24/240 & 1.04/220   & \textbf{0.94}/\textbf{208} \\ \hline
    12  & 8.44/207 & 5.08/195   & \textbf{5.07}/\textbf{186}  \\ \hline
    13  & 1.04/244 & 0.65/204   & \textbf{0.60}/\textbf{202}  \\ \hline 
    14  & 2.57/228 & \textbf{2.50}/218 & \textbf{2.50}/\textbf{210}\\ \hline 
    % 15  & 1.37& 0.52& \textbf{0.31}  \\ \hline
    \end{tabular}
    \caption{Detailed ablation studies to show the effect of the couple discriminator (CoD) and contrastive loss (CL) on \dslr{} when compared to best baseline on CARLA-64 test set.}
    \label{tab:ablation}
    \end{table}

\setlength{\tabcolsep}{1.5pt}
\begin{table}[htbp]
\scriptsize
\centering
\begin{tabular}{|c|cc|}
\hline
Dataset                   & \multicolumn{1}{c|}{\begin{tabular}[c]{@{}c@{}}Vanilla Discriminator based GAN\end{tabular}} & \dslr{}                \\ \hline
                          & \multicolumn{2}{c|}{CD/EMD}                                                                                           \\ \hline
\multirow{7}{*}{CARLA-64} & \multicolumn{1}{c|}{1.58/243.52}                                                               & \textbf{0.37/235.90} \\ \cline{2-3} 
                          & \multicolumn{1}{c|}{1.48/198.36}                                                               & \textbf{0.33/156.14} \\ \cline{2-3} 
                          & \multicolumn{1}{c|}{5.11/463.12}                                                               & \textbf{1.19/226.81} \\ \cline{2-3} 
                          & \multicolumn{1}{c|}{1.59/292.21}                                                               & \textbf{0.95/207.79} \\ \cline{2-3} 
                          & \multicolumn{1}{c|}{\textbf{3.54}/271.63}                                                               & 5.36\textbf{/186.35} \\ \cline{2-3} 
                          & \multicolumn{1}{c|}{1.71/341.26}                                                               & \textbf{0.6/202.30}  \\ \cline{2-3} 
                          & \multicolumn{1}{c|}{8.92/\textbf{185.71}}                                                               & \textbf{2.63/}210.48 \\ \hline
\end{tabular}

\caption{Ablation study to show the comparison between a vanilla discriminator based GAN and our couple discriminator based \dslr{}.}
\label{tab:ablation-disc}
\end{table}

\dslr{} has 2 modules: Couple Discriminator(CoD) and Contrastive Loss(CL). We perform ablation studies to show the effect of these modules on \dslr{}. CoD is our couple-discriminator based Relativistic Average Least Squares GAN (Column 3 of Table \ref{tab:ablation}). It bridges a significant gap between the best baseline and our model. CL coupled with CoD (Column 4) gives further improvement on Chamfer's Distance. 

We also conduct another ablation study to demonstrate that a standard vanilla discriminator based Relativistic Average Least Square GAN performs inferior to our couple discriminator based \dslr{}. (Table \ref{tab:ablation-disc}).

\section{Video Demo}

We demonstrate \dslr{} ability to generate precise static reconstructions and segment movable and moving objects through video demos. In the video demo at \cite{url1}, we visualize the results of the static reconstructions of MOVES and compare it with the best baseline. In this video demo at \cite{url2}, we compare the performance of our models on a the sparse industrial dataset - ARD-16. Our model is able to segment out the movable and moving objects effectively when compared to the best baseline. In this video demo at \cite{url3}, we visualize the segmentation of the movable and the moving objects in a LiDAR sequence.

\section{Conclusion}
In this work, we address the challenge that Autonomous Vehicles(AVs) face in the absence of labelled segmentation information. We try to close the gap between settings that have several labelled datasets available and several label-devoid settings. Non-stationary object segmentation and accurate static structures reconstruction are major research challenges in these settings. We present a novel GAN-based adversarial model, \dslr{} that is augmented with a robust pair discriminator. We utilize a smartly selected LiDAR triplet using hard-negative mining that serves two purposes - generate latent pairs for the discriminator training (and) augment the discriminator with contrastive loss to increase its discriminative capacity. Ablation studies show the benefit of this setting over a vanilla discriminator setting. \dslr{} segments out accurate \textit{movable} and moving objects using static reconstructions without using labelled information. The generated static structures that replace dynamic objects and corresponding occlusions supplement the available static structures which improves downstream task performance. Our work is of practical benefit for several industrial settings - that may not have labelled segmentation and object information due to several constraints. We also demonstrate that our method is adaptable to datasets that lack static-dynamic correspondence information and performs well in the absence of segmentation-labels. \dslr{} currently is not able to surpass the performance in the labelled settings - where labelleld segmentation information is available for navigation. We do perform better for long sequences in the presence of high dynamism and multiple loop closures. These are realistic and frequent scenarios that AVs come across. For moderate to low dynamism sequences, segmentation based methods perform better. A promising future work in this direction is to find more accurate static reconstruction methods to address this gap and perform better than baselines that use segmentation information. Another future work of immediate interest that we have planned to work on is to develop methods that can generate  paired correspondence information from dataset that do not have such correspondence pairs available (e.g. KITTI). This would enable us to eliminate the need for Domain Adaptation and boost static performance for existing unpaired datasets.

\section{Acknowledgment}
This work was supported by the Indian Institute of Technology, Delhi, India.

\bibliographystyle{elsarticle}
\bibliography{references}

\begin{thebibliography}{37}
\expandafter\ifx\csname natexlab\endcsname\relax\def\natexlab#1{#1}\fi
\providecommand{\url}[1]{\texttt{#1}}
\providecommand{\href}[2]{#2}
\providecommand{\path}[1]{#1}
\providecommand{\DOIprefix}{doi:}
\providecommand{\ArXivprefix}{arXiv:}
\providecommand{\URLprefix}{URL: }
\providecommand{\Pubmedprefix}{pmid:}
\providecommand{\doi}[1]{\href{http://dx.doi.org/#1}{\path{#1}}}
\providecommand{\Pubmed}[1]{\href{pmid:#1}{\path{#1}}}
\providecommand{\bibinfo}[2]{#2}
\ifx\xfnm\relax \def\xfnm[#1]{\unskip,\space#1}\fi
%Type = Article
\bibitem[{Chen et~al.(2021)Chen, Li, Mersch, Wiesmann, Gall, Behley, and
  Stachniss}]{chen2021moving}
\bibinfo{author}{X.~Chen}, \bibinfo{author}{S.~Li},
  \bibinfo{author}{B.~Mersch}, \bibinfo{author}{L.~Wiesmann},
  \bibinfo{author}{J.~Gall}, \bibinfo{author}{J.~Behley},
  \bibinfo{author}{C.~Stachniss},
\newblock \bibinfo{title}{Moving object segmentation in 3d lidar data: A
  learning-based approach exploiting sequential data},
\newblock \bibinfo{journal}{IEEE Robotics and Automation Letters}
  \bibinfo{volume}{6} (\bibinfo{year}{2021}) \bibinfo{pages}{6529--6536}.
%Type = Misc
\bibitem[{Kumar et~al.(2021)Kumar, Sahoo, Shah, Kondameedi, Jain, Verma,
  Bhattacharyya, and Vishwanath}]{dslr-git}
\bibinfo{author}{P.~Kumar}, \bibinfo{author}{S.~Sahoo},
  \bibinfo{author}{V.~Shah}, \bibinfo{author}{V.~Kondameedi},
  \bibinfo{author}{A.~Jain}, \bibinfo{author}{A.~Verma},
  \bibinfo{author}{C.~Bhattacharyya}, \bibinfo{author}{V.~Vishwanath},
  \bibinfo{title}{Dslr: Dynamic to static lidar scan reconstruction using
  adversarially trained autoencoder}, \bibinfo{year}{2021}. \URLprefix
  \url{https://github.com/dslrproject/dslr/tree/master}.
%Type = Article
\bibitem[{Wang et~al.(2022)Wang, Wu, and Sun}]{wang2022coarse}
\bibinfo{author}{T.~Wang}, \bibinfo{author}{L.~Wu}, \bibinfo{author}{C.~Sun},
\newblock \bibinfo{title}{A coarse-to-fine approach for dynamic-to-static image
  translation},
\newblock \bibinfo{journal}{Pattern Recognition} \bibinfo{volume}{123}
  (\bibinfo{year}{2022}) \bibinfo{pages}{108373}.
%Type = Article
\bibitem[{Ahn et~al.(2023)Ahn, Kim, Kang, Pae, and Lim}]{ahn2023unsupervised}
\bibinfo{author}{W.-J. Ahn}, \bibinfo{author}{D.-W. Kim},
  \bibinfo{author}{T.-K. Kang}, \bibinfo{author}{D.-S. Pae},
  \bibinfo{author}{M.-T. Lim},
\newblock \bibinfo{title}{Unsupervised semantic segmentation inpainting network
  using a generative adversarial network with preprocessing},
\newblock \bibinfo{journal}{Applied Sciences} \bibinfo{volume}{13}
  (\bibinfo{year}{2023}) \bibinfo{pages}{781}.
%Type = Article
\bibitem[{Fei et~al.(2023)Fei, Yang, Ma, and Chen}]{fei2023dctr}
\bibinfo{author}{B.~Fei}, \bibinfo{author}{W.~Yang}, \bibinfo{author}{L.~Ma},
  \bibinfo{author}{W.-M. Chen},
\newblock \bibinfo{title}{Dctr: Noise-robust point cloud completion by
  dual-channel transformer with cross-attention},
\newblock \bibinfo{journal}{Pattern Recognition} \bibinfo{volume}{133}
  (\bibinfo{year}{2023}) \bibinfo{pages}{109051}.
%Type = Article
\bibitem[{Li et~al.(2023)Li, Liu, Huang, Xia, Yang, and Lu}]{li2023progressive}
\bibinfo{author}{P.~Li}, \bibinfo{author}{X.~Liu}, \bibinfo{author}{J.~Huang},
  \bibinfo{author}{D.~Xia}, \bibinfo{author}{J.~Yang}, \bibinfo{author}{Z.~Lu},
\newblock \bibinfo{title}{Progressive generation of 3d point clouds with
  hierarchical consistency},
\newblock \bibinfo{journal}{Pattern Recognition} \bibinfo{volume}{136}
  (\bibinfo{year}{2023}) \bibinfo{pages}{109200}.
%Type = Article
\bibitem[{Zhang et~al.(2023)Zhang, Liu, Xie, Nie, Zhou, Tao, and
  Li}]{zhang2023learning}
\bibinfo{author}{S.~Zhang}, \bibinfo{author}{X.~Liu}, \bibinfo{author}{H.~Xie},
  \bibinfo{author}{L.~Nie}, \bibinfo{author}{H.~Zhou},
  \bibinfo{author}{D.~Tao}, \bibinfo{author}{X.~Li},
\newblock \bibinfo{title}{Learning geometric transformation for point cloud
  completion},
\newblock \bibinfo{journal}{International Journal of Computer Vision}
  (\bibinfo{year}{2023}) \bibinfo{pages}{1--21}.
%Type = Inproceedings
\bibitem[{Zhang et~al.(2021)Zhang, Chen, Cai, Pan, Zhao, Yi, Yeo, Dai, and
  Loy}]{zhang2021unsupervised}
\bibinfo{author}{J.~Zhang}, \bibinfo{author}{X.~Chen},
  \bibinfo{author}{Z.~Cai}, \bibinfo{author}{L.~Pan},
  \bibinfo{author}{H.~Zhao}, \bibinfo{author}{S.~Yi}, \bibinfo{author}{C.~K.
  Yeo}, \bibinfo{author}{B.~Dai}, \bibinfo{author}{C.~C. Loy},
\newblock \bibinfo{title}{Unsupervised 3d shape completion through gan
  inversion},
\newblock in: \bibinfo{booktitle}{Proceedings of the IEEE/CVF Conference on
  Computer Vision and Pattern Recognition}, \bibinfo{year}{2021}, pp.
  \bibinfo{pages}{1768--1777}.
%Type = Inproceedings
\bibitem[{Kumar et~al.(2021)Kumar, Sahoo, Shah, Kondameedi, Jain, Verma,
  Bhattacharyya, and Vishwanath}]{kumar2021dynamic}
\bibinfo{author}{P.~Kumar}, \bibinfo{author}{S.~Sahoo},
  \bibinfo{author}{V.~Shah}, \bibinfo{author}{V.~Kondameedi},
  \bibinfo{author}{A.~Jain}, \bibinfo{author}{A.~Verma},
  \bibinfo{author}{C.~Bhattacharyya}, \bibinfo{author}{V.~Vishwanath},
\newblock \bibinfo{title}{Dynamic to static lidar scan reconstruction using
  adversarially trained auto encoder},
\newblock in: \bibinfo{booktitle}{Proceedings of the AAAI Conference on
  Artificial Intelligence}, volume~\bibinfo{volume}{35}, \bibinfo{year}{2021},
  pp. \bibinfo{pages}{1836--1844}.
%Type = Inproceedings
\bibitem[{Yang et~al.(2021)Yang, Zou, Kong, Huang, Liu, Li, Wen, and
  Zhang}]{yang2021semantic}
\bibinfo{author}{X.~Yang}, \bibinfo{author}{H.~Zou}, \bibinfo{author}{X.~Kong},
  \bibinfo{author}{T.~Huang}, \bibinfo{author}{Y.~Liu},
  \bibinfo{author}{W.~Li}, \bibinfo{author}{F.~Wen},
  \bibinfo{author}{H.~Zhang},
\newblock \bibinfo{title}{Semantic segmentation-assisted scene completion for
  lidar point clouds},
\newblock in: \bibinfo{booktitle}{2021 IEEE/RSJ International Conference on
  Intelligent Robots and Systems (IROS)}, \bibinfo{organization}{IEEE},
  \bibinfo{year}{2021}, pp. \bibinfo{pages}{3555--3562}.
%Type = Article
\bibitem[{Rist et~al.(2021)Rist, Emmerichs, Enzweiler, and
  Gavrila}]{rist2021semantic}
\bibinfo{author}{C.~B. Rist}, \bibinfo{author}{D.~Emmerichs},
  \bibinfo{author}{M.~Enzweiler}, \bibinfo{author}{D.~M. Gavrila},
\newblock \bibinfo{title}{Semantic scene completion using local deep implicit
  functions on lidar data},
\newblock \bibinfo{journal}{IEEE transactions on pattern analysis and machine
  intelligence} \bibinfo{volume}{44} (\bibinfo{year}{2021})
  \bibinfo{pages}{7205--7218}.
%Type = Inproceedings
\bibitem[{Xia et~al.(2023)Xia, Liu, Li, Zhu, Ma, Li, Hou, and
  Qiao}]{xia2023scpnet}
\bibinfo{author}{Z.~Xia}, \bibinfo{author}{Y.~Liu}, \bibinfo{author}{X.~Li},
  \bibinfo{author}{X.~Zhu}, \bibinfo{author}{Y.~Ma}, \bibinfo{author}{Y.~Li},
  \bibinfo{author}{Y.~Hou}, \bibinfo{author}{Y.~Qiao},
\newblock \bibinfo{title}{Scpnet: Semantic scene completion on point cloud},
\newblock in: \bibinfo{booktitle}{Proceedings of the IEEE/CVF Conference on
  Computer Vision and Pattern Recognition}, \bibinfo{year}{2023}, pp.
  \bibinfo{pages}{17642--17651}.
%Type = Inproceedings
\bibitem[{Zyrianov et~al.(2022)Zyrianov, Zhu, and Wang}]{zyrianov2022learning}
\bibinfo{author}{V.~Zyrianov}, \bibinfo{author}{X.~Zhu},
  \bibinfo{author}{S.~Wang},
\newblock \bibinfo{title}{Learning to generate realistic lidar point clouds},
\newblock in: \bibinfo{booktitle}{European Conference on Computer Vision},
  \bibinfo{organization}{Springer}, \bibinfo{year}{2022}, pp.
  \bibinfo{pages}{17--35}.
%Type = Article
\bibitem[{Zhang et~al.(2023)Zhang, Zhang, Kuang, and Zhang}]{zhang2023nerf}
\bibinfo{author}{J.~Zhang}, \bibinfo{author}{F.~Zhang},
  \bibinfo{author}{S.~Kuang}, \bibinfo{author}{L.~Zhang},
\newblock \bibinfo{title}{Nerf-lidar: Generating realistic lidar point clouds
  with neural radiance fields},
\newblock \bibinfo{journal}{arXiv preprint arXiv:2304.14811}
  (\bibinfo{year}{2023}).
%Type = Inproceedings
\bibitem[{Xiong et~al.(2023)Xiong, Ma, Wang, and Urtasun}]{xiong2023learning}
\bibinfo{author}{Y.~Xiong}, \bibinfo{author}{W.-C. Ma},
  \bibinfo{author}{J.~Wang}, \bibinfo{author}{R.~Urtasun},
\newblock \bibinfo{title}{Learning compact representations for lidar completion
  and generation},
\newblock in: \bibinfo{booktitle}{Proceedings of the IEEE/CVF Conference on
  Computer Vision and Pattern Recognition}, \bibinfo{year}{2023}, pp.
  \bibinfo{pages}{1074--1083}.
%Type = Inproceedings
\bibitem[{Hadsell et~al.(2006)Hadsell, Chopra, and
  LeCun}]{hadsell2006dimensionality}
\bibinfo{author}{R.~Hadsell}, \bibinfo{author}{S.~Chopra},
  \bibinfo{author}{Y.~LeCun},
\newblock \bibinfo{title}{Dimensionality reduction by learning an invariant
  mapping},
\newblock in: \bibinfo{booktitle}{2006 IEEE Computer Society Conference on
  Computer Vision and Pattern Recognition (CVPR'06)},
  volume~\bibinfo{volume}{2}, \bibinfo{organization}{IEEE},
  \bibinfo{year}{2006}, pp. \bibinfo{pages}{1735--1742}.
%Type = Article
\bibitem[{Weinberger and Saul(2009)}]{weinberger2009distance}
\bibinfo{author}{K.~Q. Weinberger}, \bibinfo{author}{L.~K. Saul},
\newblock \bibinfo{title}{Distance metric learning for large margin nearest
  neighbor classification.},
\newblock \bibinfo{journal}{Journal of machine learning research}
  \bibinfo{volume}{10} (\bibinfo{year}{2009}).
%Type = Inproceedings
\bibitem[{Sohn(2016)}]{sohn2016improved}
\bibinfo{author}{K.~Sohn},
\newblock \bibinfo{title}{Improved deep metric learning with multi-class n-pair
  loss objective},
\newblock in: \bibinfo{booktitle}{Advances in neural information processing
  systems}, \bibinfo{year}{2016}, pp. \bibinfo{pages}{1857--1865}.
%Type = Article
\bibitem[{Robinson et~al.(2020)Robinson, Chuang, Sra, and
  Jegelka}]{robinson2020contrastive}
\bibinfo{author}{J.~Robinson}, \bibinfo{author}{C.-Y. Chuang},
  \bibinfo{author}{S.~Sra}, \bibinfo{author}{S.~Jegelka},
\newblock \bibinfo{title}{Contrastive learning with hard negative samples},
\newblock \bibinfo{journal}{arXiv preprint arXiv:2010.04592}
  (\bibinfo{year}{2020}).
%Type = Article
\bibitem[{Xie et~al.(2022)Xie, Zhan, Liu, Ong, and Loy}]{xie2022delving}
\bibinfo{author}{J.~Xie}, \bibinfo{author}{X.~Zhan}, \bibinfo{author}{Z.~Liu},
  \bibinfo{author}{Y.-S. Ong}, \bibinfo{author}{C.~C. Loy},
\newblock \bibinfo{title}{Delving into inter-image invariance for unsupervised
  visual representations},
\newblock \bibinfo{journal}{International Journal of Computer Vision}
  \bibinfo{volume}{130} (\bibinfo{year}{2022}) \bibinfo{pages}{2994--3013}.
%Type = Inproceedings
\bibitem[{Caccia et~al.(2019)Caccia, Van~Hoof, Courville, and
  Pineau}]{caccia2018deep}
\bibinfo{author}{L.~Caccia}, \bibinfo{author}{H.~Van~Hoof},
  \bibinfo{author}{A.~Courville}, \bibinfo{author}{J.~Pineau},
\newblock \bibinfo{title}{Deep generative modeling of lidar data},
\newblock in: \bibinfo{booktitle}{2019 IEEE/RSJ International Conference on
  Intelligent Robots and Systems (IROS)}, \bibinfo{organization}{IEEE},
  \bibinfo{year}{2019}, pp. \bibinfo{pages}{5034--5040}.
%Type = Article
\bibitem[{Shah et~al.(2020)Shah, Tamuly, Raghunathan, Jain, and
  Netrapalli}]{shah2020pitfalls}
\bibinfo{author}{H.~Shah}, \bibinfo{author}{K.~Tamuly},
  \bibinfo{author}{A.~Raghunathan}, \bibinfo{author}{P.~Jain},
  \bibinfo{author}{P.~Netrapalli},
\newblock \bibinfo{title}{The pitfalls of simplicity bias in neural networks},
\newblock \bibinfo{journal}{Advances in Neural Information Processing Systems}
  \bibinfo{volume}{33} (\bibinfo{year}{2020}) \bibinfo{pages}{9573--9585}.
%Type = Inproceedings
\bibitem[{Mao et~al.(2017)Mao, Li, Xie, Lau, Wang, and
  Paul~Smolley}]{mao2017least}
\bibinfo{author}{X.~Mao}, \bibinfo{author}{Q.~Li}, \bibinfo{author}{H.~Xie},
  \bibinfo{author}{R.~Y. Lau}, \bibinfo{author}{Z.~Wang},
  \bibinfo{author}{S.~Paul~Smolley},
\newblock \bibinfo{title}{Least squares generative adversarial networks},
\newblock in: \bibinfo{booktitle}{Proceedings of the IEEE international
  conference on computer vision}, \bibinfo{year}{2017}, pp.
  \bibinfo{pages}{2794--2802}.
%Type = Article
\bibitem[{Borgwardt et~al.(2006)Borgwardt, Gretton, Rasch, Kriegel,
  Sch{\"o}lkopf, and Smola}]{Borgwardt2006IntegratingSB}
\bibinfo{author}{K.~Borgwardt}, \bibinfo{author}{A.~Gretton},
  \bibinfo{author}{M.~J. Rasch}, \bibinfo{author}{H.~Kriegel},
  \bibinfo{author}{B.~Sch{\"o}lkopf}, \bibinfo{author}{A.~Smola},
\newblock \bibinfo{title}{Integrating structured biological data by kernel
  maximum mean discrepancy},
\newblock \bibinfo{journal}{Bioinformatics} \bibinfo{volume}{22 14}
  (\bibinfo{year}{2006}) \bibinfo{pages}{e49--57}.
%Type = Article
\bibitem[{Dosovitskiy et~al.(2017)Dosovitskiy, Ros, Codevilla, Lopez, and
  Koltun}]{dosovitskiy2017carla}
\bibinfo{author}{A.~Dosovitskiy}, \bibinfo{author}{G.~Ros},
  \bibinfo{author}{F.~Codevilla}, \bibinfo{author}{A.~Lopez},
  \bibinfo{author}{V.~Koltun},
\newblock \bibinfo{title}{Carla: An open urban driving simulator},
\newblock \bibinfo{journal}{arXiv preprint arXiv:1711.03938}
  (\bibinfo{year}{2017}).
%Type = Inproceedings
\bibitem[{Geiger et~al.(2012)Geiger, Lenz, and Urtasun}]{geiger2012we}
\bibinfo{author}{A.~Geiger}, \bibinfo{author}{P.~Lenz},
  \bibinfo{author}{R.~Urtasun},
\newblock \bibinfo{title}{Are we ready for autonomous driving? the kitti vision
  benchmark suite},
\newblock in: \bibinfo{booktitle}{2012 IEEE Conference on Computer Vision and
  Pattern Recognition}, \bibinfo{organization}{IEEE}, \bibinfo{year}{2012}, pp.
  \bibinfo{pages}{3354--3361}.
%Type = Inproceedings
\bibitem[{Dosovitskiy et~al.(2017)Dosovitskiy, Ros, Codevilla, Lopez, and
  Koltun}]{Dosovitskiy17}
\bibinfo{author}{A.~Dosovitskiy}, \bibinfo{author}{G.~Ros},
  \bibinfo{author}{F.~Codevilla}, \bibinfo{author}{A.~Lopez},
  \bibinfo{author}{V.~Koltun},
\newblock \bibinfo{title}{{CARLA}: {An} open urban driving simulator},
\newblock in: \bibinfo{booktitle}{Proceedings of the 1st Annual Conference on
  Robot Learning}, \bibinfo{year}{2017}, pp. \bibinfo{pages}{1--16}.
%Type = Inproceedings
\bibitem[{Hess et~al.(2016)Hess, Kohler, Rapp, and Andor}]{hess2016real}
\bibinfo{author}{W.~Hess}, \bibinfo{author}{D.~Kohler},
  \bibinfo{author}{H.~Rapp}, \bibinfo{author}{D.~Andor},
\newblock \bibinfo{title}{Real-time loop closure in 2d lidar slam},
\newblock in: \bibinfo{booktitle}{2016 IEEE International Conference on
  Robotics and Automation (ICRA)}, \bibinfo{organization}{IEEE},
  \bibinfo{year}{2016}, pp. \bibinfo{pages}{1271--1278}.
%Type = Inproceedings
\bibitem[{Bescos et~al.(2019)Bescos, Neira, Siegwart, and
  Cadena}]{bescos2019empty}
\bibinfo{author}{B.~Bescos}, \bibinfo{author}{J.~Neira},
  \bibinfo{author}{R.~Siegwart}, \bibinfo{author}{C.~Cadena},
\newblock \bibinfo{title}{Empty cities: Image inpainting for a
  dynamic-object-invariant space},
\newblock in: \bibinfo{booktitle}{2019 International Conference on Robotics and
  Automation (ICRA)}, \bibinfo{organization}{IEEE}, \bibinfo{year}{2019}, pp.
  \bibinfo{pages}{5460--5466}.
%Type = Inproceedings
\bibitem[{Fan et~al.(2017)Fan, Su, and Guibas}]{fan2017point}
\bibinfo{author}{H.~Fan}, \bibinfo{author}{H.~Su}, \bibinfo{author}{L.~J.
  Guibas},
\newblock \bibinfo{title}{A point set generation network for 3d object
  reconstruction from a single image},
\newblock in: \bibinfo{booktitle}{Proceedings of the IEEE conference on
  computer vision and pattern recognition}, \bibinfo{year}{2017}, pp.
  \bibinfo{pages}{605--613}.
%Type = Misc
\bibitem[{Hao-Su(2017)}]{emd-cd}
\bibinfo{author}{Hao-Su}, \bibinfo{title}{3d deep learning on point cloud
  representation (analysis)}, \bibinfo{year}{2017}. \URLprefix
  \url{http://graphics.stanford.edu/courses/cs468-17-spring/LectureSlides/L14%20-%203d%20deep%20learning%20on%20point%20cloud%20representation%20(analysis).pdf}.
%Type = Inproceedings
\bibitem[{Kang et~al.(2014)Kang, Ye, Li, and Doermann}]{kang2014convolutional}
\bibinfo{author}{L.~Kang}, \bibinfo{author}{P.~Ye}, \bibinfo{author}{Y.~Li},
  \bibinfo{author}{D.~Doermann},
\newblock \bibinfo{title}{Convolutional neural networks for no-reference image
  quality assessment},
\newblock in: \bibinfo{booktitle}{Proceedings of the IEEE conference on
  computer vision and pattern recognition}, \bibinfo{year}{2014}, pp.
  \bibinfo{pages}{1733--1740}.
%Type = Article
\bibitem[{Horn(1987)}]{horn1987closed}
\bibinfo{author}{B.~K. Horn},
\newblock \bibinfo{title}{Closed-form solution of absolute orientation using
  unit quaternions},
\newblock \bibinfo{journal}{Josa a} \bibinfo{volume}{4} (\bibinfo{year}{1987})
  \bibinfo{pages}{629--642}.
%Type = Inproceedings
\bibitem[{Sturm et~al.(2012)Sturm, Engelhard, Endres, Burgard, and
  Cremers}]{sturm2012benchmark}
\bibinfo{author}{J.~Sturm}, \bibinfo{author}{N.~Engelhard},
  \bibinfo{author}{F.~Endres}, \bibinfo{author}{W.~Burgard},
  \bibinfo{author}{D.~Cremers},
\newblock \bibinfo{title}{A benchmark for the evaluation of rgb-d slam
  systems},
\newblock in: \bibinfo{booktitle}{2012 IEEE/RSJ International Conference on
  Intelligent Robots and Systems}, \bibinfo{organization}{IEEE},
  \bibinfo{year}{2012}, pp. \bibinfo{pages}{573--580}.
%Type = Misc
\bibitem[{Google(2023{\natexlab{a}})}]{url1}
\bibinfo{author}{Google}, \bibinfo{title}{Static translation comparsion},
  \bibinfo{year}{2023}{\natexlab{a}}. \URLprefix
  \url{https://drive.google.com/file/d/1YDX4PBMQJMbuelSR_Ox7XjzXuOWu6bu_/view?usp=sharing}.
%Type = Misc
\bibitem[{Google(2023{\natexlab{b}})}]{url2}
\bibinfo{author}{Google}, \bibinfo{title}{Ard16 movable comparsion},
  \bibinfo{year}{2023}{\natexlab{b}}. \URLprefix
  \url{https://drive.google.com/file/d/1TrjQSsl4muwoww4bSs3CdWeb3DrsgCli/view?usp=sharing}.
%Type = Misc
\bibitem[{Google(2023{\natexlab{c}})}]{url3}
\bibinfo{author}{Google}, \bibinfo{title}{Lidar moving and movable},
  \bibinfo{year}{2023}{\natexlab{c}}. \URLprefix
  \url{https://drive.google.com/file/d/1tP4tP2EnOftkCFxeX6JZbBZB6FCjd0ow/view?usp=sharing}.

\end{thebibliography}

\end{document}